\documentclass{article}

\usepackage{microtype}
\usepackage{graphicx}
\usepackage{subfigure}
\usepackage{booktabs} 
\include{m.symbols}
\usepackage{amsfonts,amsmath,amsthm} 
\usepackage{bm,amsbsy}
\usepackage{hyperref}

\usepackage{bbm}
\usepackage{tikz}
\usepackage{siunitx}

\usepackage[preprint]{icml2019}

\icmltitlerunning{Interpretable continuous-time latent stochastic dynamical models}

\begin{document}

\twocolumn[
\icmltitle{Learning interpretable continuous-time models\\ of latent stochastic dynamical systems}







\icmlsetsymbol{equal}{*}

\begin{icmlauthorlist}
\icmlauthor{Lea Duncker}{gatsby}
\icmlauthor{Gerg\H{o} Bohner}{gatsby}
\icmlauthor{Julien Boussard}{stanford}
\icmlauthor{Maneesh Sahani}{gatsby}
\end{icmlauthorlist}

\icmlaffiliation{gatsby}{Gatsby Computational Neuroscience Unit, University College London, London, United Kingdom}
\icmlaffiliation{stanford}{Stanford University, Palo Alto, California, USA}

\icmlcorrespondingauthor{Lea Duncker}{duncker@gatsby.ucl.ac.uk}

\icmlkeywords{Gaussian Process, Stochastic Differential Equation, Variational Inference, inducing points, dynamics, fixed points}

\vskip 0.3in
]



\printAffiliationsAndNotice{}  

\begin{abstract}
%
We develop an approach to learn an interpretable semi-parametric model of a latent continuous-time stochastic dynamical
system, assuming noisy high-dimensional outputs sampled at uneven times. The dynamics are described by a nonlinear
stochastic differential equation (SDE) driven by a Wiener process, with a drift evolution function drawn from a Gaussian
process (GP) conditioned on a set of learnt fixed points and corresponding local Jacobian matrices.  This form yields a
flexible nonparametric model of the dynamics, with a representation corresponding directly to the interpretable
portraits routinely employed in the study of nonlinear dynamical systems.  The learning algorithm combines inference of
continuous latent paths underlying observed data with a sparse variational description of the dynamical process. We
demonstrate our approach on simulated data from different nonlinear dynamical systems.
\end{abstract}

\section{Introduction}
\label{sec:intro}
A wide range of dynamical systems with intrinsic noise may be modelled in continuous time using the framework of
stochastic differential equations (SDE).  However identifying a good SDE model from intermittent observations of the
process is challenging, particularly if the dynamical process is nonlinear and the observations are indirect and noisy.
A common response is to assume a latent process that operates in discretised time, often called a state-space model.
This approach has been applied in contexts ranging from modelling human motion \cite{wang+al:2006:nips} to solving
control problems \cite{eleftheriadis+al:2017:nips}.  However, it assumes that observations, and the critical phenomena
of the dynamics, can be accurately modelled using a discrete time grid.

A further challenge when the goal is to gain insight into a physical or biological system whose parametric description
is unknown, is to obtain an \emph{interpretable} model of the dynamics from observed data, whether modelled in discrete
or continuous time.  State-space models that rely on nonparametric or flexibly parametrised descriptions of dynamics,
for example using Gaussian process (GP) priors or recurrent neural networks (RNN), may be effective at prediction but
inevitably leave interpretation to a second analytic stage, posing its own challenges.

%

In this paper, we consider continuous-time latent SDE models of the form
\begin{equation}
    \begin{aligned}
       & d\vec x = \vec f(\vec x) dt + \sqrt{\vec \Sigma}\; d \vec w \\
       & \mathbb{E}_{y|x}[\vec y(t_i)] =   g(\vec C \vec x(t_i) +\vec d)  \; , \quad\quad\quad i = 1,\dots, T \,,
    \end{aligned}
\label{eq:generative_model}
\end{equation}
where the temporal evolution of a latent variable $\vec x \in \mathbb{R}^K$ is described by a nonlinear SDE with
dynamical evolution function $\vec f: \mathbb{R}^K \mapsto \mathbb{R}^K$ and incremental noise covariance $\vec \Sigma$
shaping the Wiener noise process $\vec w(t)$.  Note that the nonlinear SDE induces a non-Gaussian prior on $\vec x(t)$
with no easy access to finite marginal distributions.  The latent state is observed indirectly through noisy
measurements $\vec y_i \in \mathbb{R}^N$ at unevenly spaced time points $t_i$. The measurements are distributed with a
known parametric form and generalized linear dependence; that is the expected value is $g(\vec C \vec x +\vec d)$ with
inverse-link function $g$ and parameters $\vec C \in \mathbb{R}^{N \times K}$ and $\vec d \in \mathbb{R}^N$.  We seek to
infer latent paths $\vec x(t)$ along with the dynamical parameters and an interpretable representation of the dynamical
mapping $\vec f$.

What do we mean by interpretable?  The properties of dynamical systems are frequently analyzed by characterizing
dynamical fixed points and local behaviour near these points \cite{sussillo+al:2013:neuralcomp}.  When $\vec f$ is a
learnt, general function, fixed points must be found numerically \cite{golub+al:2019:joss}. 
%
%
This makes it difficult to propagate uncertainty about $\vec f$ to the number and location of fixed points, and to the local dynamics around them.
Our approach is to develop a
non-parametric Gaussian-process model for $\vec f$ \emph{conditioned on} the learnt locations of fixed points and
associated local Jacobians. Thus, we implicitly integrate out the details of $\vec f$, while optimising directly over
the components of the intepretable dynamical portrait.

%
%

 The paper is organised as follows: In section \ref{sec:background} we will review background material on the related Gaussian Process State-Space Model (GP-SSM) as well as previous work on Gaussian Process approximations to SDEs \cite{archambeau+al:2007:jmlr, archambeau+al:2008:nips}. We will also briefly review the inducing point approach for Gaussian Process models. In section \ref{sec:GP_param} we will make use of Gaussian Process priors to represent the unknown nonlinear dynamics $\vec f$, incorporating interpretable structure by conditioning the Gaussian Process on fixed points and local Jacobian matrices of the system. We derive a Variational Bayes algorithm for approximate inference and parameter learning in section \ref{sec:inference_learning}, allowing for efficient closed form updates. Finally, we demonstrate the performance of our alogrithm on a number of nonlinear dynamical system examples in section \ref{sec:results}.

\section{Background}
\label{sec:background}

\subsection{Gaussian Process State-Space-Model}
A discrete-time analogue of the model in (\ref{eq:generative_model}) is the Gaussian Process State-Space Model (GP-SSM), where the latent state evolution over a fixed step size is modelled as
\begin{equation}
    \vec x_{\ell+1} = \vec f (\vec x_\ell) + \vec \epsilon_\ell
    \label{eq:gpssm}
\end{equation}
where $\vec \epsilon_\ell \sim \mathcal{N}(\vec \epsilon_\ell |0, D)$.
There have been a range of approaches for performing approximate inference in this model, based on Assumed Density Filtering \cite{deisenroth+al:2009:icml,ramakrishnan+al:2011:ieee}, Expectation Propagation \cite{deisenroth+al:2012:nips}, variational inference \cite{frigola+al:2014:nips}, or recurrent recognition networks \cite{eleftheriadis+al:2017:nips}. The model we consider in this paper requires a different treatment for latent path inference, as it maintains the continuous-time structure of the system of interest. 

\subsection{Gaussian Process Approximation to SDEs}

 The problem of performing approximate inference in continuous-time SDE models has been considered previously, with the two main approaches being Expectation Propagation \cite{cseke+al:2016:journalofphysics} and variational inference \cite{archambeau+al:2007:jmlr, archambeau+al:2008:nips}. We will review the latter approach in this section, as our Variational Bayes algorithm in section \ref{sec:inference_learning} will extend this work.
 
\citet{archambeau+al:2007:jmlr, archambeau+al:2008:nips} consider the model in (\ref{eq:generative_model}) under linear Gaussian observations. The authors derive an approximate inference algorithm based on a variational Gaussian approximation to the posterior process on $\vec x(t)$ under the constraint that the approximate process has Markov structure, as is the case for the true posterior process. The most general way to construct such an approximation is via a linear time-varying SDE of the form
\begin{equation}
    d \vec x = \left (-\vec A(t) \vec x(t) + \vec b(t) \right) dt + \sqrt{\vec \Sigma} \; d\vec w
    \label{eq:q_sde}
\end{equation}

An alternative way to express this approximation is via a GP of the form
\begin{equation}
    q_x(\vec x(t)) =\mathcal{GP}\left (\vec m_x(t),\vec S_x(t)\right) 
\label{eq:q_marginal}
\end{equation}
whose means $\vec m_x(t)$ and covariances $\vec S_x(t)$ evolve in time according to the ordinary differential equations (ODEs):
\begin{equation}
    \begin{aligned}
    \frac{d \vec m_x}{dt} & = - \vec A(t) \vec m_x  + \vec b(t)\\
    \frac{d \vec S_x}{dt} & =  - \vec A(t) \vec S_x - \vec S_x \vec A(t) \tr + \vec \Sigma \\
    \end{aligned}
    \label{eq:q_odes}
\end{equation}

\citet{archambeau+al:2007:jmlr, archambeau+al:2008:nips} derive a lower bound to the marginal log-likelihood -- often called the variational free energy or evidence lower bound -- whose maximisation with respect to $q_x$ is equivalent to minimising the Kullback-Leibler (KL) divergence between the approximate and true posterior process. The free energy has the form
\begin{equation}
    \mathcal{F} = \sum_i \left\langle \log p(\vec y_i | \vec x_i)\right \rangle_{q_x} - \KL{q_x(\vec x)}{p(\vec x)}
\label{eq:free_energy}
\end{equation}
The first term is the expected log-likelihood under the approximation and only depends on the marginal distributions $q_x(\vec x(t))$. The second term is the KL-divergence between the continuous-time approximate posterior process and the prior process. \citet{archambeau+al:2007:jmlr} show that this term can be written as 
\begin{equation}
    \KL{q_x(\vec x)}{p(\vec x)} = \int_\mathcal{T} dt  \left\langle (\vec f - \vec f_q )\tr \vec \Sigma\inv (\vec f - \vec f_q) \right\rangle_q 
\end{equation}
where both $\vec f$ and $\vec f_q$ are evaluated at $\vec x(t)$ though not explicitly written, and $\vec f_q(\vec x(t)) = - \vec A(t) \vec x(t) + \vec b(t)$. Note that the noise covariance $\vec \Sigma$ is deliberately chosen to be equal for the SDEs in $q_x$ and $p$, as this term would diverge otherwise.

To maximise $\mathcal{F}$ with respect to $\vec m_x(t)$ and $\vec S_x(t)$, subject to the constraint that the approximate posterior process has Markov structure according to equation (\ref{eq:q_sde}), one can find the stationary points of the Lagrangian
\begin{equation}
         \mathcal{L} = \mathcal{F} - \mathcal{C}_1 - \mathcal{C}_2
\label{eq:Lagrangian}
\end{equation}
with 
\begin{equation}
\begin{aligned}
    \mathcal{C}_1 & = \int_{\mathcal{T}}dt\; \trace{\Psi(\frac{d\vec S_x}{dt} + \vec A \vec S_x + \vec S_x \vec A\tr -\vec \Sigma)} \\
    \mathcal{C}_2 & = \int_{\mathcal{T}}dt\; \vec \lambda\tr (\frac{d\vec m_x}{dt} +\vec A \vec m_x -\vec b)
\end{aligned}
\label{eq:Lagrangian_constraints}
\end{equation}

Where $\Psi$ and $\vec \lambda$ are Lagrange multipliers. \citet{archambeau+al:2007:jmlr,archambeau+al:2008:nips} derive a smoothing algorithm that involves iterating fixed point updates of this Lagrangian. These are either closed form, or require solving ODEs forward and backward in time, thus achieving linear time complexity. In section \ref{sec:inference_learning}, we will modify this original algorithm in order to improve its numerical stability, and show how to incorporate it in an efficient Variational Bayes algorithm. 

\subsection{Sparse Gaussian Processes using inducing points}

In later sections of the paper, we will make use of the sparse variational inducing point approach of \citet{titsias+al:2009:aistats}. The key idea of inducing point approaches is to condition a GP $\zeta(\vec x) \sim \mathcal{GP}(0, \kappa(\vec x,\vec x'))$ on what can be thought of as pseudo-observations of the function at $M$ locations $\vec Z = [\vec z_1, \dots, \vec z_M] \in \mathbb{R}^{K \times M}$. These pseudo-observation are the inducing points $\vec u \in \mathbb{R}^{M}$. An augmented prior for the GP and inducing-points can be written as
\begin{equation}
    \begin{aligned}
    \vec u &\sim \mathcal{N}\left( \vec u| 0, \vec K_{zz}\right) \\
    \zeta | \vec u_{k} &\sim \mathcal{GP}(\mu_{\zeta|u}(\vec x), \nu_{\zeta|u}(\vec x, \vec x')
    \end{aligned}
    \label{eq:conditioned_gp}
\end{equation}
The mean and covariance function of the conditioned GP are given by
\begin{equation}
    \begin{aligned}
    \mu_{\zeta|u}(\vec x) & = \vec \kappa(\vec x, \vec Z) \vec K_{zz}\inv \vec u \\
     \nu_{\zeta|u}(\vec x, \vec x') & = \kappa(\vec x,\vec x') - \vec \kappa(\vec x, \vec Z) \vec K_{zz}\inv  \vec \kappa(\vec Z, \vec x')
    \end{aligned}
\label{eq:conditioned_gp_mean_variance}
\end{equation}
Where $[\vec K_{zz}]_{ij} = \kappa(\vec z_i, \vec z_j)$, and $[\vec \kappa(\vec x, \vec Z)]_i = \kappa(\vec x, \vec z_i)$. The computational complexity of building the mean and covariance in (\ref{eq:conditioned_gp_mean_variance}) is linear in the number of $\vec x$ input points and cubic only in the number of inducing points $M$. If we were to integrate over the inducing points in this augmented prior, we would recover the original model. However, the inducing points can also be kept in the model as auxiliary variables, which may be incorporated into approaches for variational inference \cite{titsias+al:2009:aistats}.

\section{Interpretable priors on nonlinear dynamics}
\label{sec:GP_param}

Similarly to the GP-SSM work, we wish to model $\vec f$ using the framework of GPs. GPs can represent a flexible class of nonlinear dynamics. However, it may be difficult to interpret the inferred function with respect to studying the underlying dynamical system that generated the observed data. As stated above, standard analysis approaches for nonlinear dynamical systems rely on identifying local fixed points $\vec s_i$, where $\vec f(\vec s_i) = \vec 0$, and the locally-linearised dynamics around them, given by the Jacobians $\vec \nabla_x \vec f(\vec x)|_{\vec x = \vec s_i}$ \cite{sussillo+al:2013:neuralcomp}.  This strategy motivates our approach to interpretability. 

\subsection{A Gaussian Process prior for dynamics}
In order to arrive at a modelling framework that makes fixed points and Jacobian matrices readily available for analysis, we introduce a GP prior conditioned directly on these parameters, as we have done previously in \citet{bohner+al:2018:arxiv}. 
The fixed point locations and Jacobians around them can be viewed as further hyperparameters specifying the prior mean and covariance function of the GP, which we will denote by $\vec \theta = \{\vec f_s^{(i)} ,\vec J_s^{(i)} \}_{i=1}^L$. With $\vec f_s^{(i)} = \vec f(\vec s_i) = \vec 0$ and $[\vec J_s^{(i)}]_{k,m} = \frac{\partial f_k(\vec x)}{\partial x_m}|_{\vec x = \vec s_i}$. We can hence write a GP prior conditioned on the fixed points and Jacobians for each dimension in $\vec f$, using the fact that a GP and its derivative process are still jointly distributed as a GP.

The Variational Bayes approach in section \ref{sec:inference_learning} will make use of a sparse variational approximation for $\vec f$ using inducing points, as in \citet{titsias+al:2009:aistats}. To make later notation more compact, we therefore directly introduce the augmented model including inducing points drawn from the conditioned GP prior here. 
We denote the joint covariance matrix between inducing points, fixed points and Jacobian matrices as
\begin{equation}
     \mat K_{zz}^\theta = \begin{bmatrix}
     \vec K_{zz} & \vec K_{zs} & \vec K_{zs}^{\nabla_2}\\
    \vec K_{sz} &   \vec K_{ss} & \vec K_{ss}^{\nabla_2}\\
    \vec K_{sz}^{\nabla_1} & \vec K_{ss}^{\nabla_1} &  \vec K_{ss}^{\nabla_1\nabla_2}
    \end{bmatrix} = \begin{bmatrix}
     \vec K_{zz} & \tilde{\vec K}_{zs} \\
    \tilde{\vec K}_{sz} &   \tilde{\vec K}_{ss}
    \end{bmatrix}
    \label{eq:block_covariance_matrix}
\end{equation}
where the superscript $\nabla_i$ denotes the derivative of the covariance function with respect to its $i$th input argument such that $[\vec K_{zs}^{\nabla_2}]_{ij} = \frac{\partial}{\partial \vec s} \kappa(\vec z_i, \vec s) |_{\vec s = \vec s_j}$. The conditional prior on the inducing points given $\vec \theta$ can then be written as
\begin{equation}
    \begin{aligned}
    \vec u_k | \vec \theta = \mathcal{N}\left( \vec u \Big|\; \tilde{\vec K}_{zs} {\tilde{\vec K}_{ss}}\inv \vec v_k^\theta, \vec K_{zz} - \tilde{\vec K}_{zs}  {\tilde{\vec K}_{ss}}\inv  \tilde{\vec K}_{sz}\right)    
    \end{aligned}
\end{equation}
where $\vec v_k^\theta = [ f_{s,k}^{(1)}, \dots,  f_{s,k}^{(L)},  \vec J^{(1)}_{k,:}, \dots, \vec J^{(L)}_{k,:}]\tr$ collects the fixed-point and derivative observations relating to $f_k$.
Finally, for the conditional prior on $f_k$, given the inducing points and $\vec \theta$, we have
\begin{equation}
    \begin{aligned}
    f_k | \vec u_k, \vec \theta & \sim \mathcal{GP}\left(\mu_{f|u}^{\theta}(\vec x), \nu_{f|u}^{\theta}(\vec x, \vec x')\right)
    \end{aligned}
\end{equation}
with
\begin{equation}
    \begin{aligned}
    \mu_{f|u}^{\theta}(\vec x) & = \vec a_z^\theta (\vec x) \begin{bmatrix}
    \vec u_k\\
    \vec v_k^\theta
    \end{bmatrix}\\
    \nu_{f|u}^{\theta}(\vec x, \vec x') & =  \kappa(\vec x, \vec x') -     \vec a_z^\theta (\vec x) 
    {\mat K_{zz}^\theta} \vec a_z^\theta (\vec x)\tr 
    \end{aligned}
    \label{eq:fixedPoint_priorGP}
\end{equation}
where we have defined
\begin{equation}
\vec a_z^\theta (\vec x) =  \begin{bmatrix}
    \vec \kappa(\vec x, \vec Z)  & \vec{\kappa}(\vec x, \vec S) & \nabla_2 {\vec{\kappa}}(\vec x, \vec S)
    \end{bmatrix}
    {\mat K_{zz}^\theta}\inv
\end{equation}

\subsection{Automatic selection of the number of fixed points} \label{sec:ard_fxdpts}
When the generative SDE dynamics are unknown, so are the number of fixed points in the system. We therefore take the general approach of introducing more fixed points that expected, and `pruning' by hyperparameter optimisation.  In particular, we include noise variance parameters for each fixed-point, representing uncertainty about the zero-value of the function at the fixed point location.
We hence have
\begin{equation}
    \vec f_i^{s} = \vec f(\vec s_i) + \alpha_i \vec \epsilon = \vec 0 + \alpha_i \vec \epsilon
\end{equation}
with $\vec \epsilon \sim \mathcal{N}(0,I)$. The variance parameters $\alpha_i$ will enter our model simply via an added diagonal matrix to the $\vec K_{ss}$ block in (\ref{eq:block_covariance_matrix}). When the $\alpha_i$ are optimised, the uncertainty for superfluous fixed points will grow, while that of the fixed points the system is actually using will shrink.  When the uncertainty for a fixed point is large, conditioning on it in the GP prior for $\vec f$ will essentially have no effect on prediction. 

\section{Variational inference and learning}
\label{sec:inference_learning}
We can derive an efficient Variational Bayes (VB) algorithm \cite{attias+al:2000:nips} for variational inference and learning in the model in (\ref{eq:generative_model}) by maximising a variational free energy. 
We assume that our full variational distribution factorises as 
\begin{equation}
    q(\vec x,\vec f,\vec u) =  q_x(\vec x) q_{f,u} (\vec f,\vec u )
    \label{eq:factored_var_approx}
\end{equation}
Following \citet{titsias+al:2009:aistats}, we choose $q_{f,u}(\vec f, \vec u) = \prod_{k=1}^K p(f_k | \vec u_k, \vec \theta) q_{u}(\vec u_k)$. The variational approximation of the posterior over the inducing points are chosen to be of the form $q_u(\vec u_k) = \mathcal{N}\left(\vec u_k | \vec m_{u}^k , \vec S_{u}^k \right)$. The marginal variational distribution $q_f(\vec f) = \prod_k \int d\vec u_k p(f_k | \vec u_k, \vec \theta) q_{u}(\vec u_k)$ is also a Gaussian Process.
The resulting expression for the variational free energy is of the form:
\begin{equation}
\begin{aligned}
    \mathcal{F}^{*} & = \left\langle \mathcal{F} \right\rangle_{q_{f}} - \sum_{k=1}^K \KL{q_u(\vec u_k)}{p(\vec u_k| \vec \theta)}
    \label{eq:free_energy_full}
\end{aligned}
\end{equation}

The VB algorithm then iterates over an inference step, where the distribution $q_x$ over the latent path is updated, a learning step where $q_{f,u}$ and the parameters in the affine output mapping are updated, and a hyperparameter learning step where the kernel hyperparameters, and fixed point locations are updated. 
\subsection{Inference}
Our inference approach follows directly from the work in \citet{archambeau+al:2007:jmlr, archambeau+al:2008:nips}, though we consider a wider class of observation models and include a nonparametric Bayesian treatment of the dynamics $\vec f$ under the conditioned sparse GP prior introduced in section \ref{sec:GP_param}. 

After using integration by parts on the Lagrangian in (\ref{eq:Lagrangian}) (exchanging $\mathcal{F}$ for $\mathcal{F}^*$), we take variational derivatives with respect to $\vec m_x(t)$ and $\vec S_x(t)$.  Since our model has a rotational non-identifiability with respect to the latents $\vec x$, we fix $\vec \Sigma=I$ without loss of generality. We arrive at the following set of fixed point equations:
\begin{align}
    \frac{d \vec \Psi}{dt} & = \vec A(t)\tr \vec \Psi(t) - \vec \Psi(t)\vec A(t) - \frac{\partial \mathcal{F}^*}{\partial \vec S_x} \odot \mathbb{P}  \label{eq:psi_ode}\\ 
    \frac{d \vec \lambda}{dt} & = \vec A^T(t) \vec \lambda(t) - \frac{\partial \mathcal{F^*} }{\partial \vec m_x}\label{eq:lambda_ode}\\
    \vec A(t) & = \left\langle \frac{\partial \vec f}{\partial \vec x} \right\rangle_{q_x q_f} + 2 \vec\Psi(t) \label{eq:A_update}\\
    \vec b(t) & = \left\langle \vec f(\vec x) \right\rangle_{q_x q_f} + \vec A(t) \vec m_x(t) - \vec\lambda(t) \label{eq:b_update}
\end{align}
with  ${\mathbb{P}}_{ij} = \frac{1}{2}$ for $i \ne j$ and $1$ otherwise and $\odot$ denotes the Hadamard product. In contrast to previous work, we explicitly take the symmetric variations of $\vec S_x(t)$ into account, which leads to slightly modified equations in (\ref{eq:psi_ode}) compared to the work in \citet{archambeau+al:2007:jmlr, archambeau+al:2008:nips}, and seems to improve the numerical stability of the algorithm. As a result, we can work with the fixed point updates (\ref{eq:A_update}) and (\ref{eq:b_update}) directly, without introducing a learning rate parameter that blends the updates with the previous value of the variational parameters $\vec A$ and $\vec b$, as was done in \citet{archambeau+al:2007:jmlr, archambeau+al:2008:nips}.

The inference algorithm involves solving the set of coupled ODEs in (\ref{eq:q_odes}) and (\ref{eq:psi_ode})-(\ref{eq:b_update})
using the conditions $\vec m_x(0) = \vec m_{x,0}$, $\vec S_x(0) = \vec S_{x,0}$ and $\vec \lambda(T) = 0$, $\vec \Psi(T) = 0$. In principle, it is possible to use any ODE solver to do this. In this work, we choose to solve (\ref{eq:q_odes}) using the forward Euler method with fixed step size $\Delta t$ to obtain $\vec m_x$ and $\vec S_x$ evaluated on an evenly spaced grid. Similarly, we then solve (\ref{eq:psi_ode}) and (\ref{eq:lambda_ode}) backwards in time to obtain evaluations of  $\vec \lambda$ and $\vec \Psi$. The solutions from the ODEs can then be used with equations (\ref{eq:A_update}) and (\ref{eq:b_update}) to obtain evaluations of $\vec A$ and $\vec b$ on the same time-grid used for solving the ODEs.

%

Evaluating the expectations of the terms involving $\vec f$ with respect to $q_x$ and $q_f$ only involves computing Gaussian expectations of covariance functions and their derivatives. These can be computed analytically for choices such as an exponentiated quadratic covariance function. We update the initial state values $\vec m_{x,0}$ and $\vec S_{x,0}$ using the same procedure as that described in \citet{archambeau+al:2008:nips}. Given the function evaluations on the inference time-grid, we use linear interpolation to obtain function evaluations of $\vec m_x$ and $\vec S_x$ at arbitrary time points. Further details on the inference algorithm are given in the supplementary material. 

\subsection{Learning}
\subsubsection{Dynamics}
The only terms in (\ref{eq:free_energy_full}) that depend on parameters in $\vec f$ are the expected KL-divergence between the prior and approximate posterior processes and the KL-divergence relating to the inducing points for $\vec f$, which are jointly quadratic in the inducing points and Jacobians. Thus, given $\vec m_x(t)$, $\vec S_x(t)$, $\vec A(t)$ and $\vec b(t)$, we can find closed form updates for the Jacobians and variational parameters relating to $\vec f$. For $\vec S_u^k$ the update is of the form
\begin{align}
    \vec S_u^k & = \left( {\vec \Omega_u} \inv + \int_\mathcal{T} dt  [ \langle \vec a_z^\theta (\vec x)\tr \vec a_z^\theta (\vec x) \rangle_{q_x}]_{[u,u]} \right) \inv 
\end{align}
with $\vec \Omega_u  = \vec K_{zz} - \tilde{\vec K}_{zs}  {\tilde{\vec K}_{ss}}\inv  \tilde{\vec K}_{sz}$ and where the operation $[X]_{[u,u]}$ selects the first $M\times M$ block of X.
The inducing points and Jacobians around the fixed-point locations can be updated jointly as
\begin{align*}
     \begin{bmatrix}
\vec m_u^1&  \dots &\vec m_u^K \\
\vec J_1 & \dots & \vec J_K
\end{bmatrix} & = \vec B_{1} \inv \left(\vec B_{2} -\vec B_{3} \right)
\end{align*}
with
\begin{align*}
    \vec B_{1} & = \left(\tilde{\vec \Omega} + \int_\mathcal{T} dt \left[ \langle\vec a_z^\theta (\vec x)\tr \vec a_z^\theta (\vec x) \rangle_{q_x} \right]_{[uj,uj]} \right) \\
    \vec B_{2} & =    \int_\mathcal{T} dt  \left[\left\langle \vec a_z^\theta (\vec x)\right\rangle_{q_x}\right]_{[:,uj]} \tr \langle \vec f_q \rangle_{q_x}\tr \\
     \vec B_{3} & =    \int_\mathcal{T} dt \left[\left\langle \nabla_x \vec a_z^\theta (\vec x)\right\rangle_{q_x} \right]_{[:,uj]} \tr \vec S_x \vec A \tr\\
      \tilde{\vec \Omega}  & = \begin{bmatrix}
{\vec \Omega_u}\inv & - {\vec \Omega_u}\inv \vec G\\
- \vec G\tr {\vec \Omega_u}\inv & \vec G \tr {\vec \Omega_u}\inv \vec G
\end{bmatrix},  \vec G = \left[\tilde{\vec K}_{zs} {\tilde{\vec K}_{ss}}\inv\right]_{[j,j]}
\end{align*}
 where $[X]_{[uj,uj]}$ selects the first $M\times M$ and last $LK \times LK$ block of $X$, $[X]_{[:,uj]}$ selects the first $M$ and last $LK$ columns of $X$, and  $[X]_{[j,j]}$ selects the last $LK \times LK$ block of $X$. The one-dimensional integrals can be computed efficiently using, for instance, Gauss-Legendre quadrature. Detailed derivations are given in the supplementary material, where we also provide closed form updates for the sparse variational GP approach for modelling $\vec f$ without further conditioning on fixed points and Jacobians.
 

\subsubsection{Output Mapping}
The only term that depends on the parameters $\vec C$ and $\vec d$ in (\ref{eq:free_energy_full}) is the expected log-likelihood. Whether or not our algorithm admits for closed form solutions depends on the choice of observation model. In the case of a Gaussian likelihood, we can find the optimal updates as
\begin{align}
     & \vec C^{*}  =  \left( \sum_t (\vec y_t - \vec d)\vec m_{x,t} \tr \right) \left( \sum_t (\vec S_{x,t} + \vec m_{x,t} \vec m_{x,t} \tr) \right) \inv \\
     & \vec d^{*} = \frac{1}{T}\sum_t \left( \vec y_t - \vec C^{*} \vec m_{x,t} \right)
\end{align}
Where the subscript $t$ denotes a function evaluation at $t$. For other choices of observation model a closed form solution may not be available, but parameter updates can again be found by maximising the free energy using standard optimisation approaches.

\subsubsection{Hyperparameters}
The covariance function hyperparameters and fixed point locations are learnt by direct optimisation of the variational free energy. The inducing point locations can also be included here, though we chose to hold them fixed on a chosen grid for all examples shown in this paper.

\section{Experiments}
\label{sec:results}
In this section, we apply our algorithm to data generated from different nonlinear dynamical systems. In all experiments, we choose an exponentiated quadratic covariance function in the prior over the dynamics $\vec f$ and initialise the inducing point means and Jacobian matrices at zero. Each fixed point observation's uncertainty is initialised with a standard deviation of $0.1$. We generate $\vec C$ and $\vec d$ by drawing their entries from Gaussian distributions unless otherwise stated, and initialise our algorithm at these parameter values. For inference, we solve the ODEs (\ref{eq:psi_ode})-(\ref{eq:b_update}) using the forward Euler method with $\Delta t= 1$ms.
\subsection{Double-well dynamics}
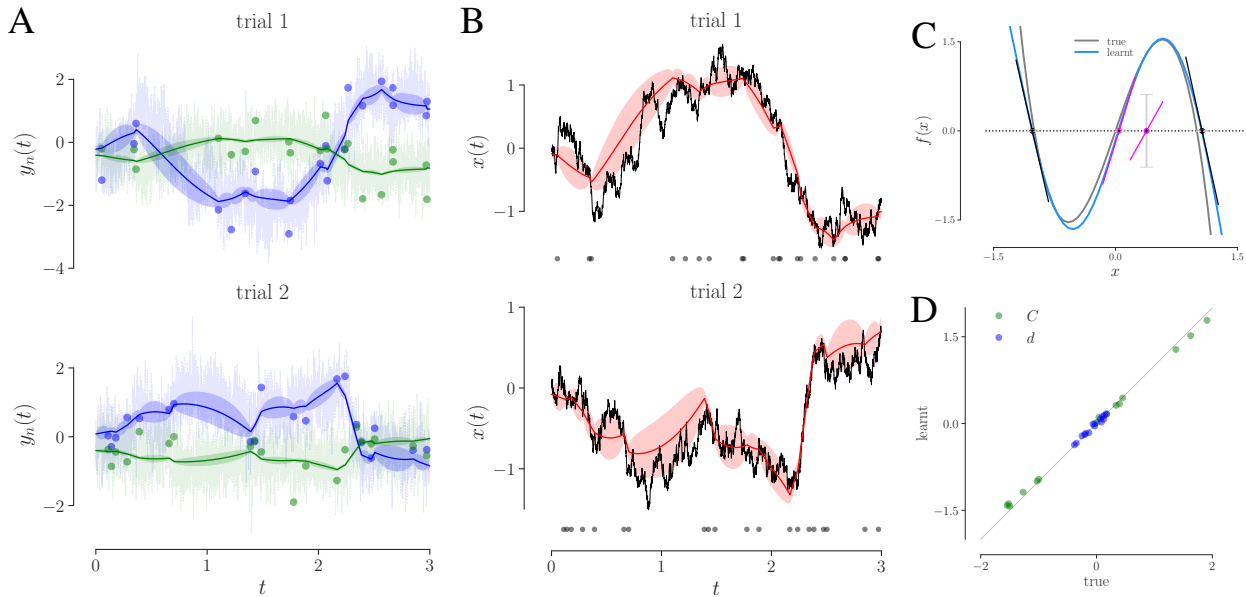
\begin{figure*}[h!]
\centering
\begin{tikzpicture}
    \node[anchor=west] (out1) at (-8,2) {\includegraphics[width=0.35\linewidth]{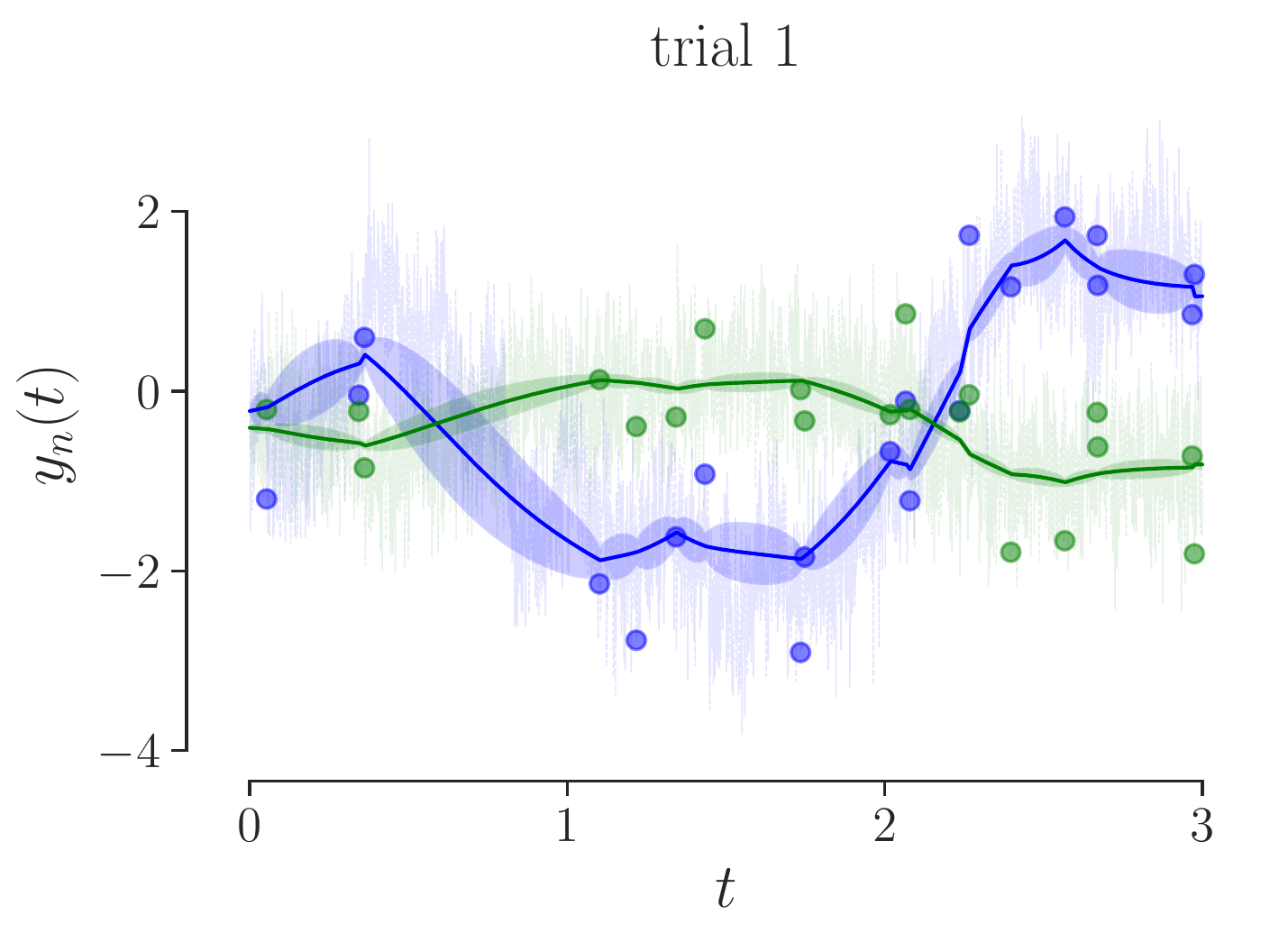}};
    \node[anchor=west] (out2) at (-8,-1.6) {\includegraphics[width=0.35\linewidth]{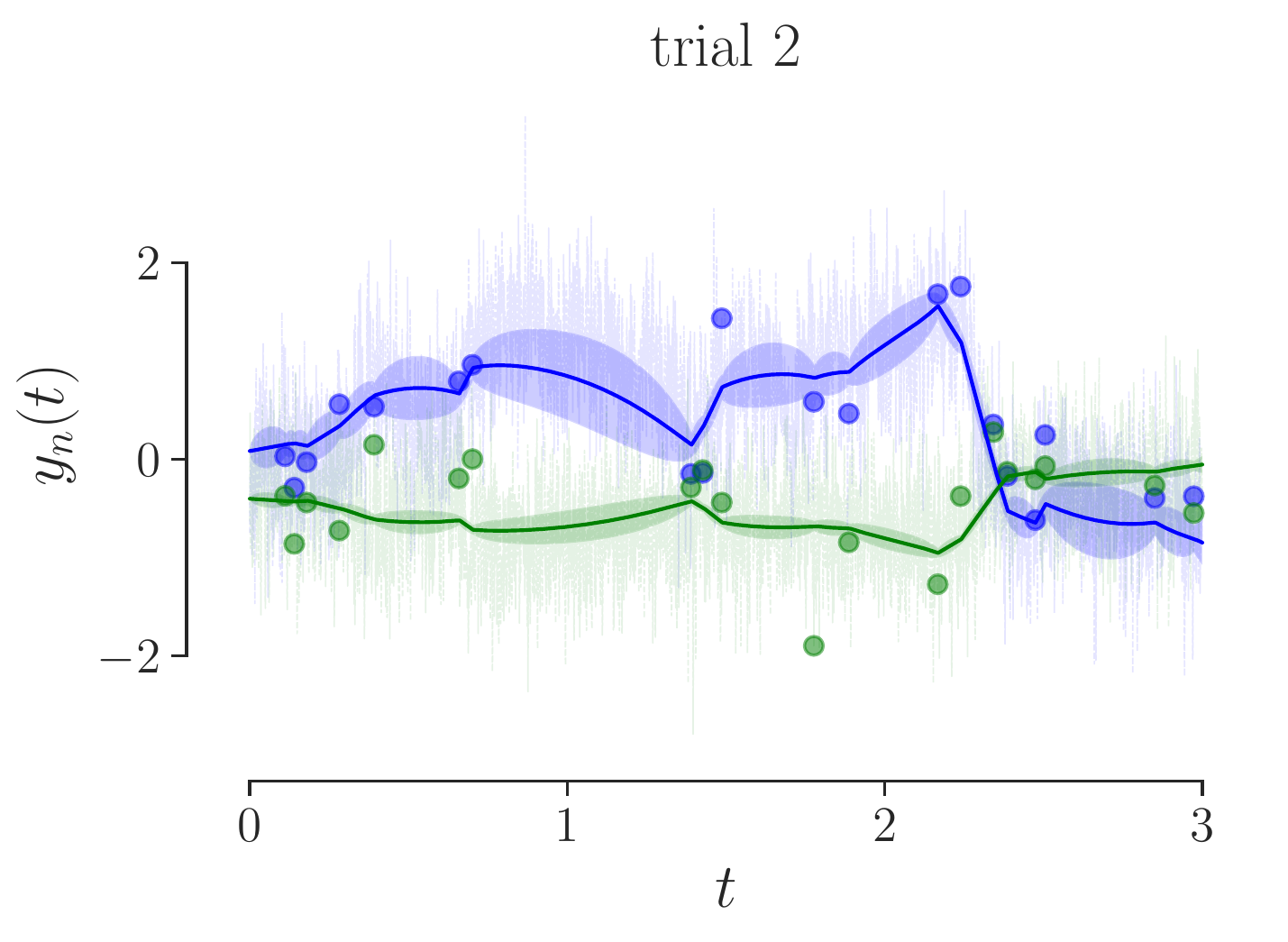}};
    \node at ([yshift=2cm,xshift=0.3cm]out1.west){\Large{A}};
    \node[anchor=west] (latent) at (-2,2) {\includegraphics[width=0.35\linewidth]{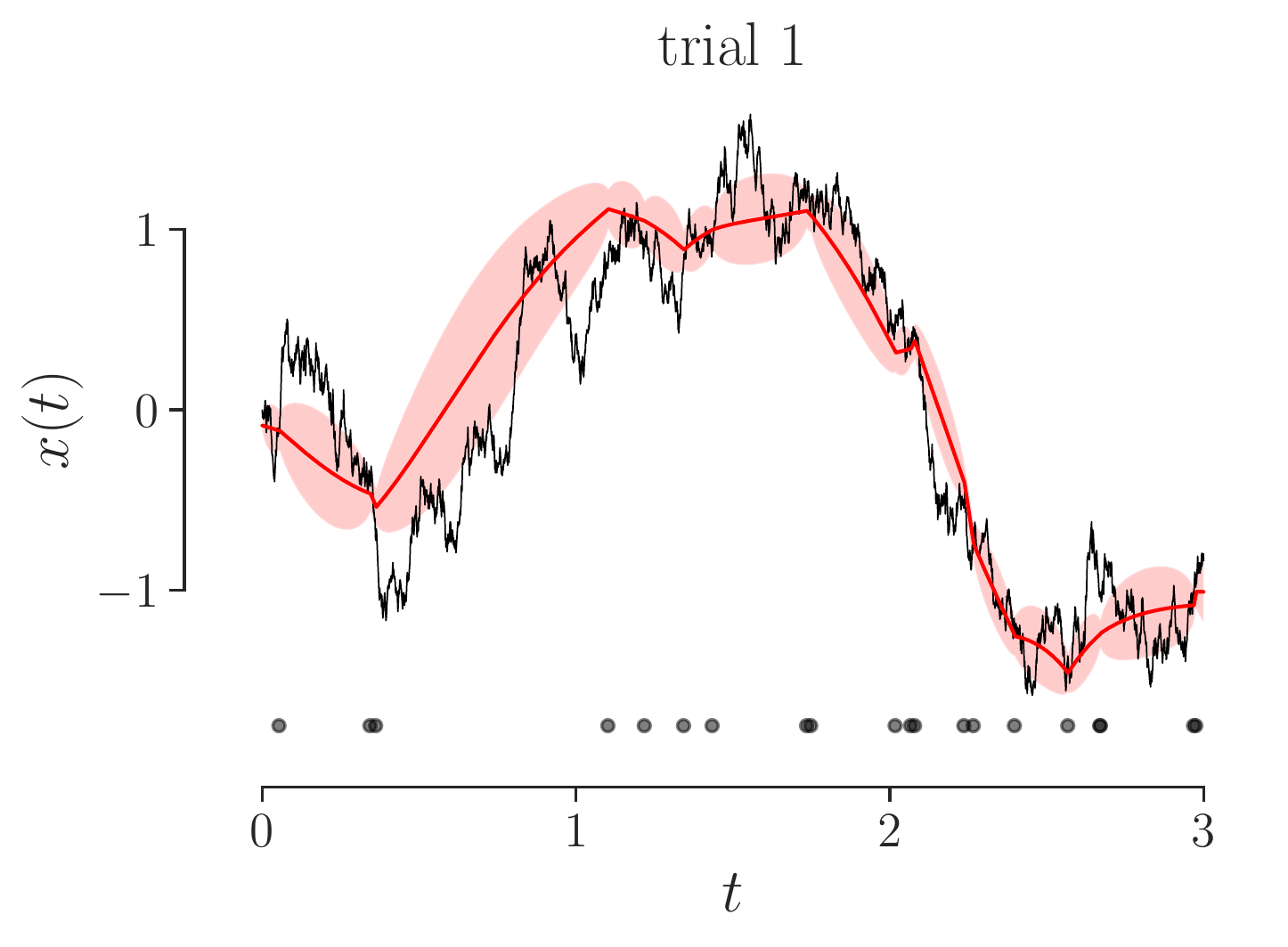}};
    \node at ([yshift=2cm,xshift=0.3cm]latent.west){\Large{B}};
    \node[anchor=west] (em) at (-2,-1.6) {\includegraphics[width=0.35\linewidth]{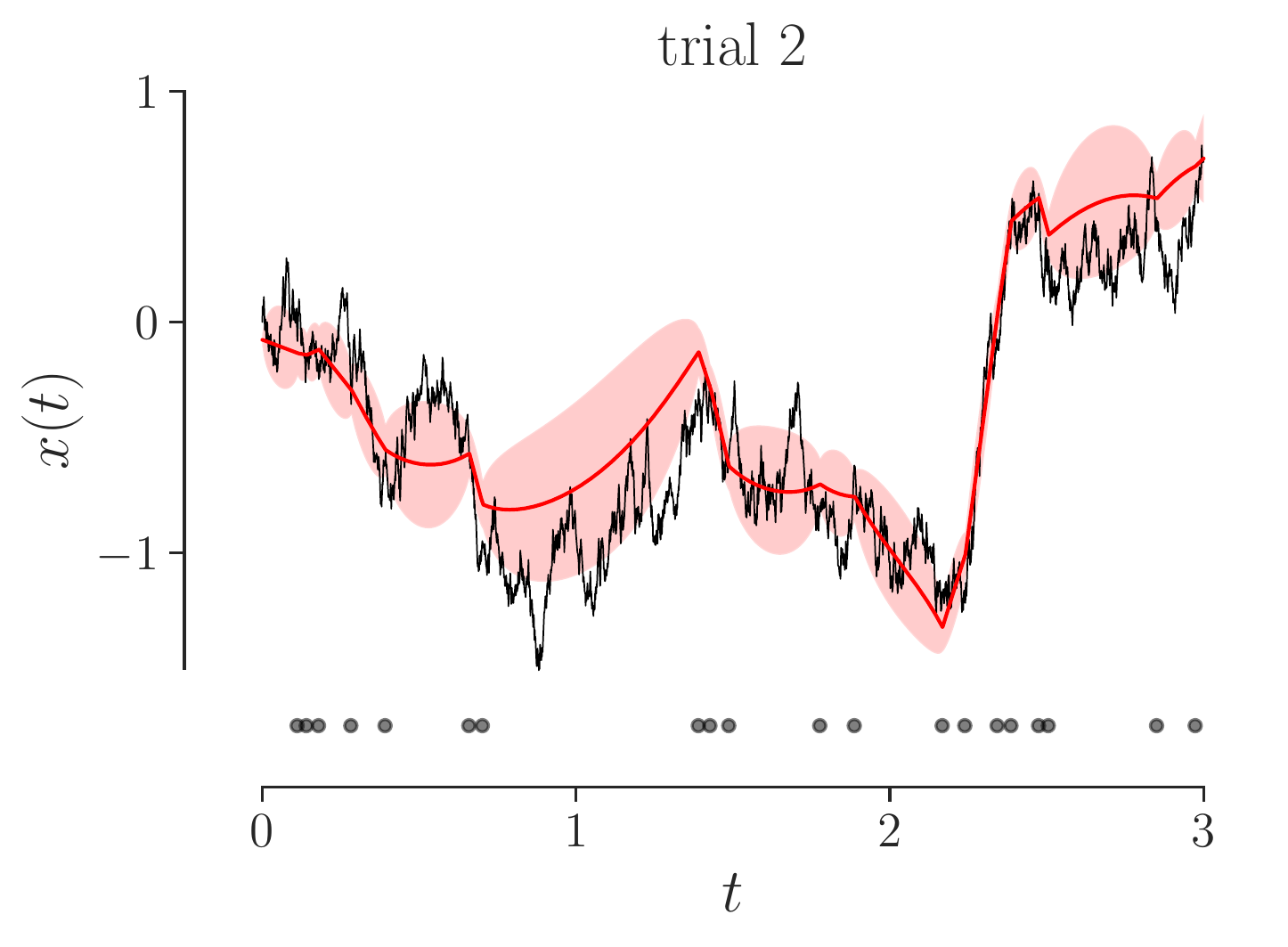}};
    \node[anchor=west] (dyn) at (4,2.25) {\includegraphics[width=0.275\linewidth]{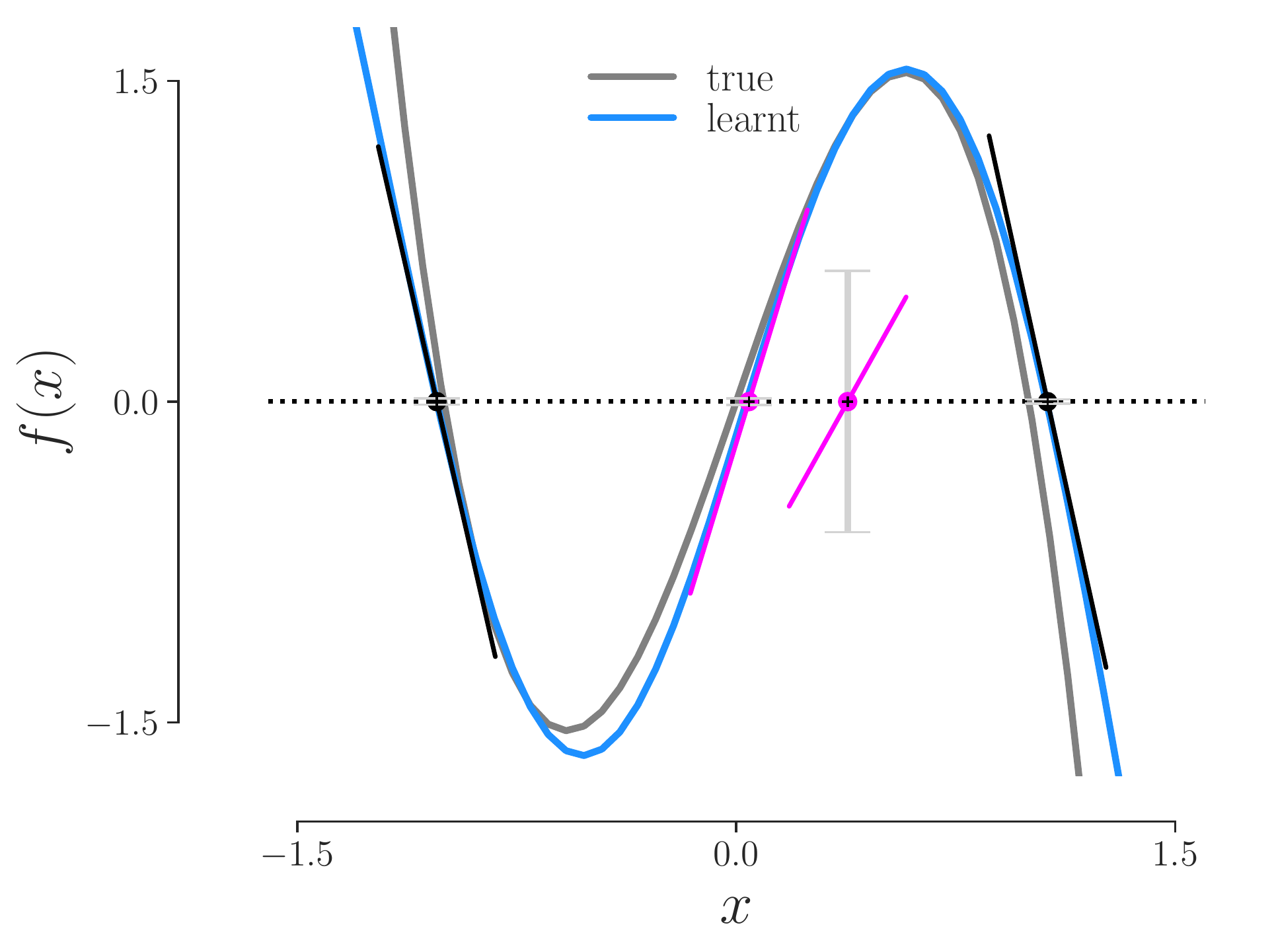}};
    \node at ([yshift=1.55cm,xshift=0.3cm]dyn.west){\Large{C}};
    \node[anchor=west] (dyn) at (4,-1.6) {\includegraphics[width=0.25\linewidth]{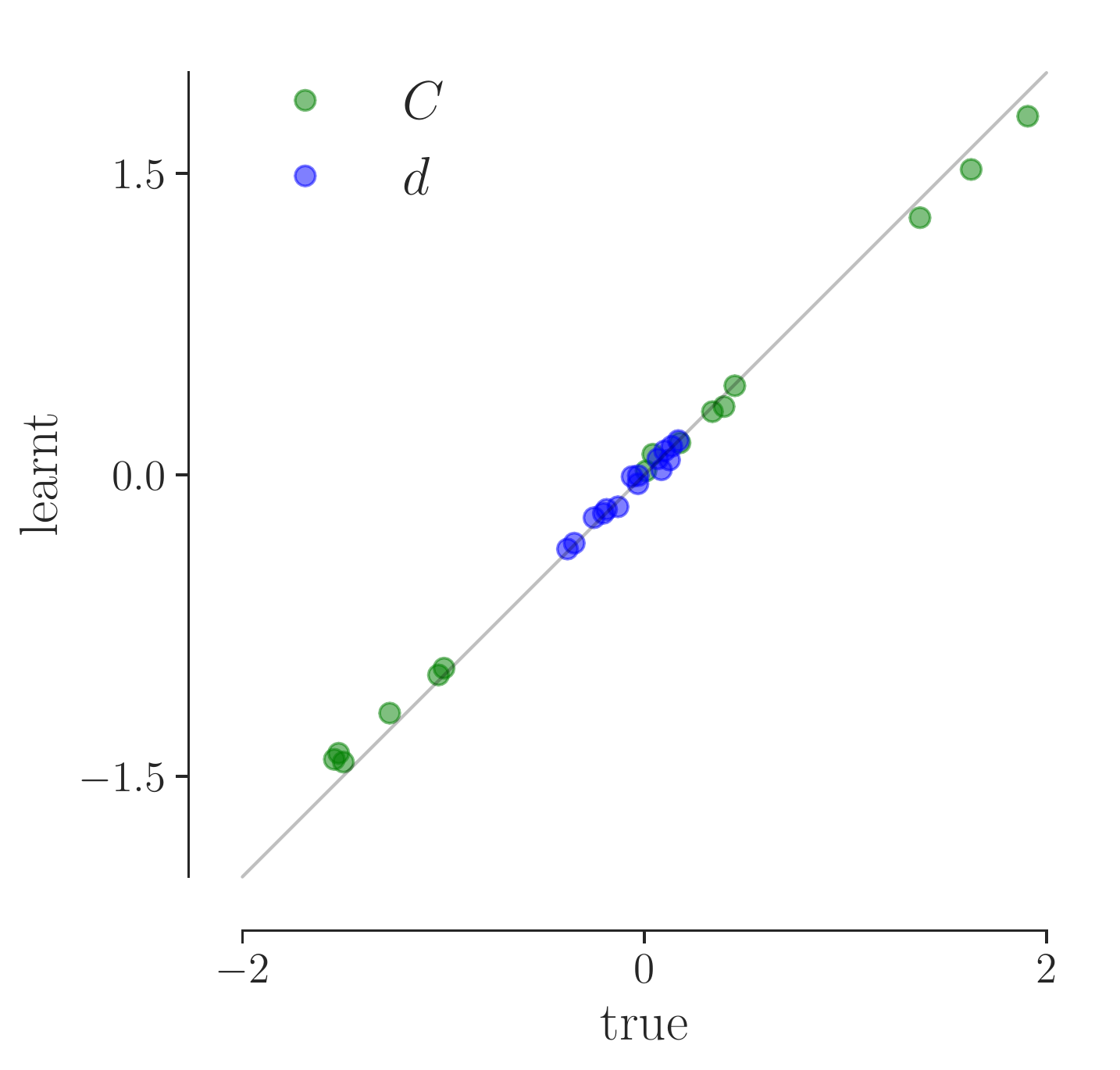}};
    \node at ([yshift=1.75cm,xshift=0.3cm]dyn.west){\Large{D}};
\end{tikzpicture}
\caption{Double-well dynamics. A: Two example dimensions of the output process on two different trials. The dots represent the observed data-points of the noisy output processes plotted in faint lines. The solid blue/green traces are the inferred posterior means with $\pm 1$ posterior standard deviation tubes around them. B: True and inferred latent SDE trajectory for the same example trials as in A. The red traces represent the posterior means with $\pm 1$ posterior standard deviation tubes around them, black traces show the true latent SDE path. The black dots indicate the times when observations of $\vec y$ were made. C: True and learnt dynamics together with the learnt fixed-point locations and tangent lines. Stable fixed points are shown in black, unstable ones in magenta. The uncertainty about the fixed point observation is illustrated using grey error bars representing $\pm 1$ standard deviation. Only the additional fourth fixed point is associated with high uncertainty. D: True vs. learnt model parameters $\vec C$ and $\vec d$.}
\label{fig:doubleWell}
\end{figure*}
We first demonstrate our method on a the classic one-dimensional double-well example, where the latent SDE evolves with drift $f(x) = 4 x (1 - x^2)$. We simulate data on 20 trials with multivariate Gaussian outputs of dimensionality $N = 15$ with unknown variances $0.25$, and observe the output process at 20 randomly sampled time-points per trial. We chose 8 evenly spaced inducing points in $(-3,3)$ for $f$. While the true dynamics have three fixed points, we condition the prior on $f$ on four fixed points and use the method outlined in section \ref{sec:ard_fxdpts} to automatically select the correct number. 
 The results are summarised in Figure \ref{fig:doubleWell}, demonstrating that our algorithm can successfully perform inference and interpretable learning of the SDE path and dynamics, respectively, and does not move away from the good initial location for the model parameters $\vec C$ and $\vec d$.

\subsection{Van der Pol's oscillator}
\begin{figure}[h!]
    \centering
    \begin{tikzpicture}
         \node[anchor=west] (dyn) at (-8.6,0) {\includegraphics[width=0.475\linewidth]{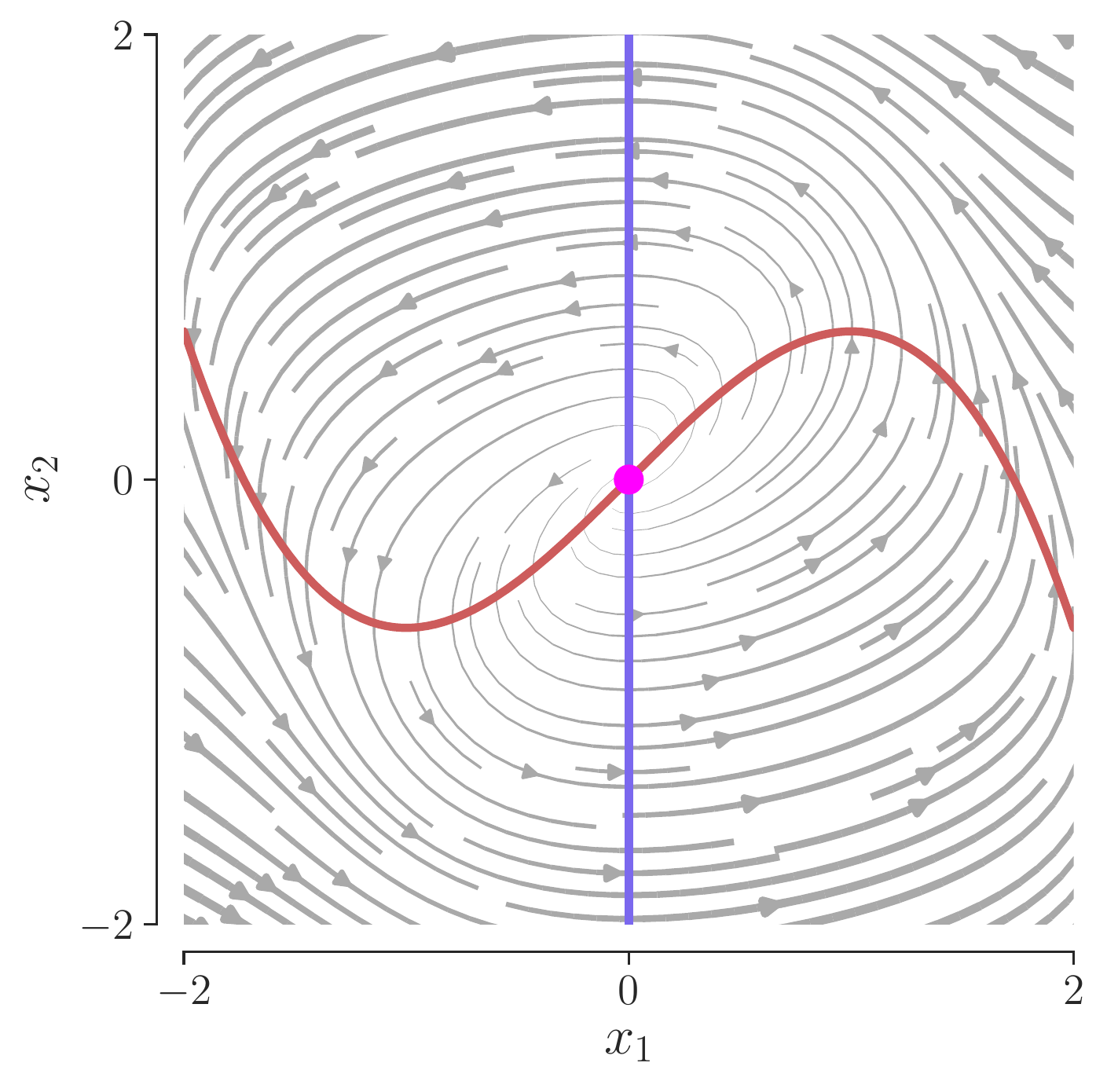}};
         \node at ([yshift=1.6cm,xshift=0.2cm]dyn.west){\Large{A}};
        \node[anchor=west] (est) at (-4.75,0) {\includegraphics[width=0.475\linewidth]{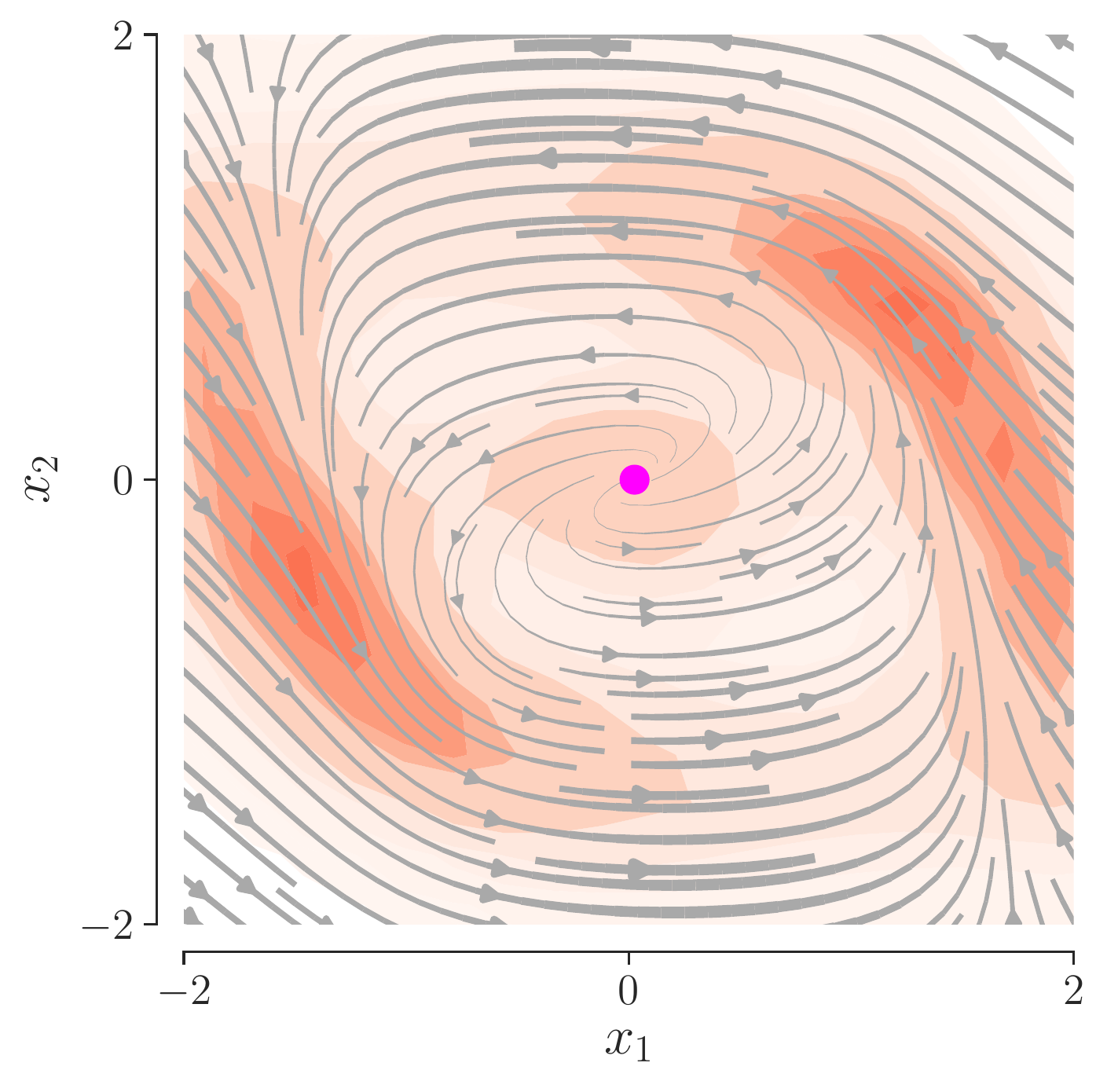}};
         \node at ([yshift=1.6cm,xshift=0.2cm]est.west){\Large{B}};
          \node[anchor=west] (out) at (-8.5,-3.7) {\includegraphics[width=0.9\linewidth]{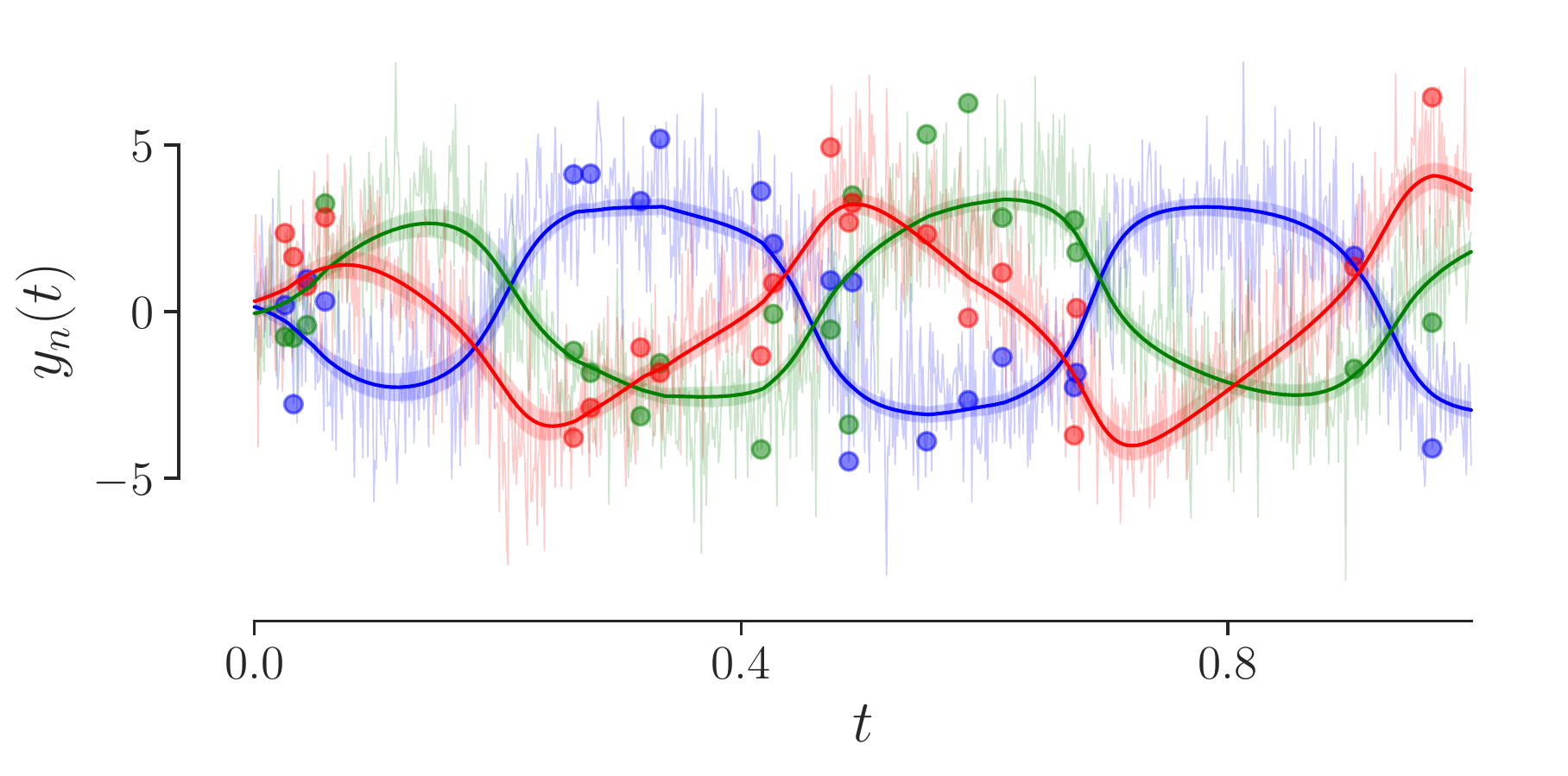}};
         \node at ([yshift=1.6cm,xshift=0.1cm]out.west){\Large{C}};
          \node[anchor=west] (out) at (-8.6,-7.1) {\includegraphics[width=0.9\linewidth]{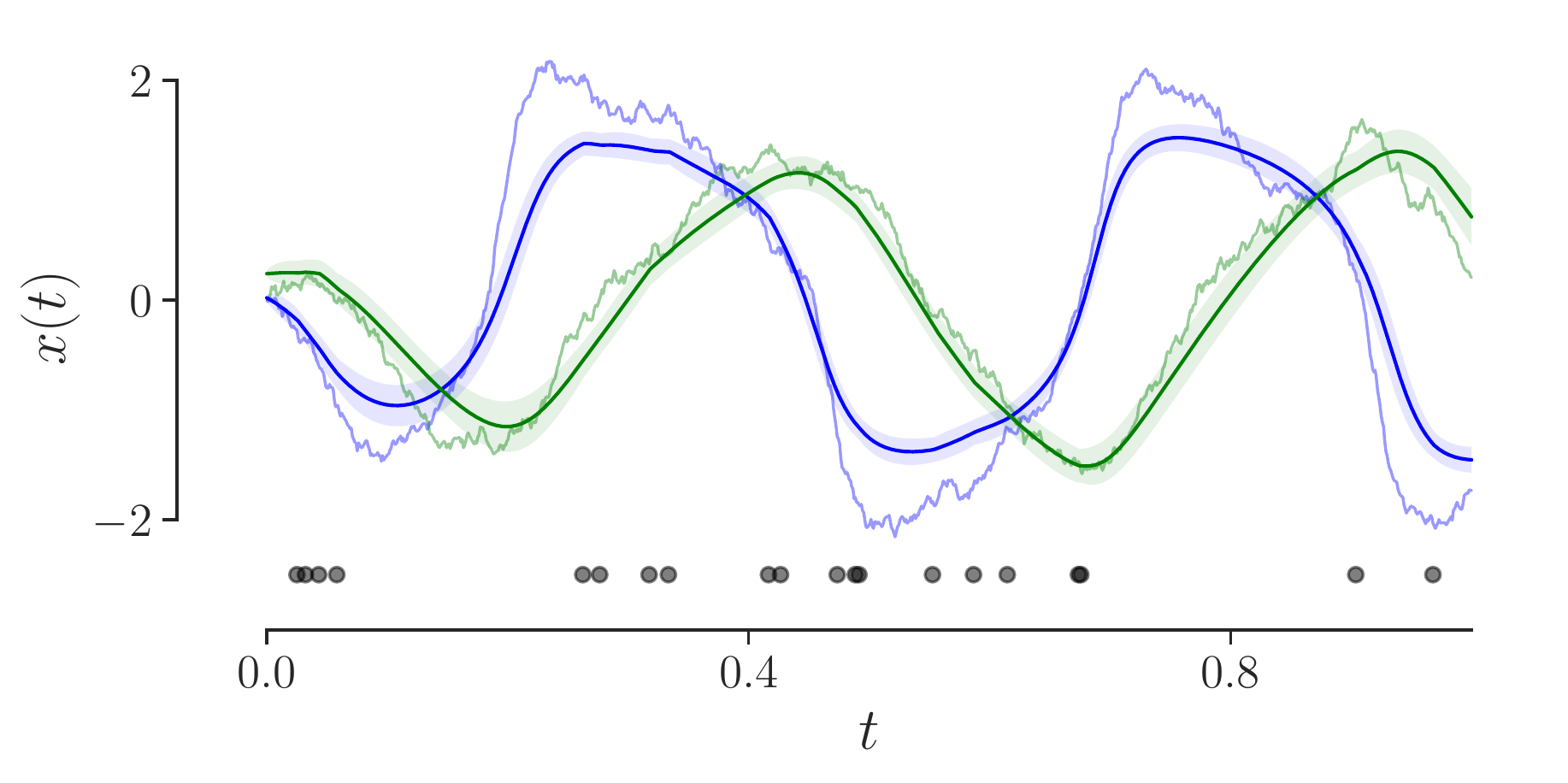}};
         \node at ([yshift=1.6cm,xshift=0.3cm]out.west){\Large{D}};
    \end{tikzpicture}
    \caption{Van der Pol's oscillator. A: Streamline plot of the true dynamics together with the nullclines and the unstable fixed point. B: Density plot of the locations visited by the latents in across all trails used for learning in red, and streamline plot of the learnt dynamics with the location of the learnt fixed point. The eigenvalues of the learnt Jacobian matrix indicate that the fixed point is unstable. C: Three example dimensions of the output process. The dots represent the observed data-points of the noisy output process. The solid traces show the the posterior means with $\pm$ 1 standard deviation tubes around them. D: The true latent SDE path together with the posterior mean $\pm$ 1 posterior standard deviation of each latent dimension. Black dots represent the locations where the 20 measurements of the output process were made.
    }
    \label{fig:vanderpols}
\end{figure}

Our next example examines a two-dimensional system where the dynamics contain a limit cycle around an unstable fixed point. The dynamics are given by
\begin{equation}
    \begin{aligned}
    f_1(\vec x) & = \rho \tau \left(x_1 - \frac{1}{3} x_1^3 - x_2\right),  && f_2( \vec x) & = \frac{\tau}{\rho} x_1
    \end{aligned}
\end{equation}
with a time constant $\tau$. We generate data from (\ref{eq:generative_model}) using these dynamics with $\rho=2, \tau=15$, $N=20$ output dimensions and Gaussian measurement noise with unknown variances $2.25$ on 20 repeated trials. We use $5\times5$ inducing points evenly spaced in $(-2,2)$. The results are summarised in Figure \ref{fig:vanderpols}, demonstrating that our description of the dynamics successfully captures the limit cycle of the generative dynamics.

\subsection{Neural population dynamics}
\begin{figure}[h!]
\centering
\begin{tikzpicture}
    \node[anchor=west] (dyn1) at (-8,2.5) {\includegraphics[width=0.495\linewidth]{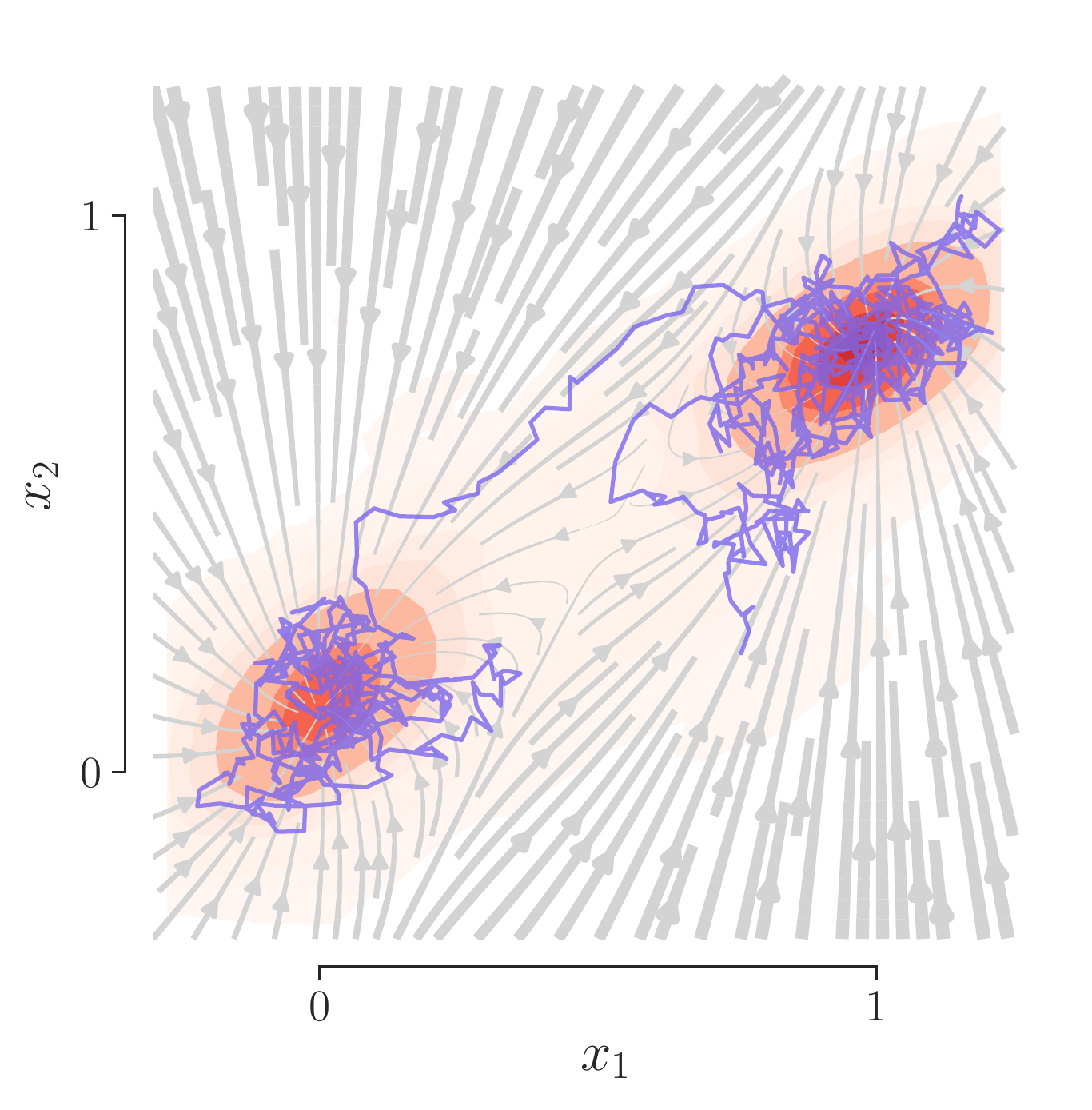}};
    \node[anchor=west] (dyn2) at (-4.1,2.5) {\includegraphics[width=0.495\linewidth]{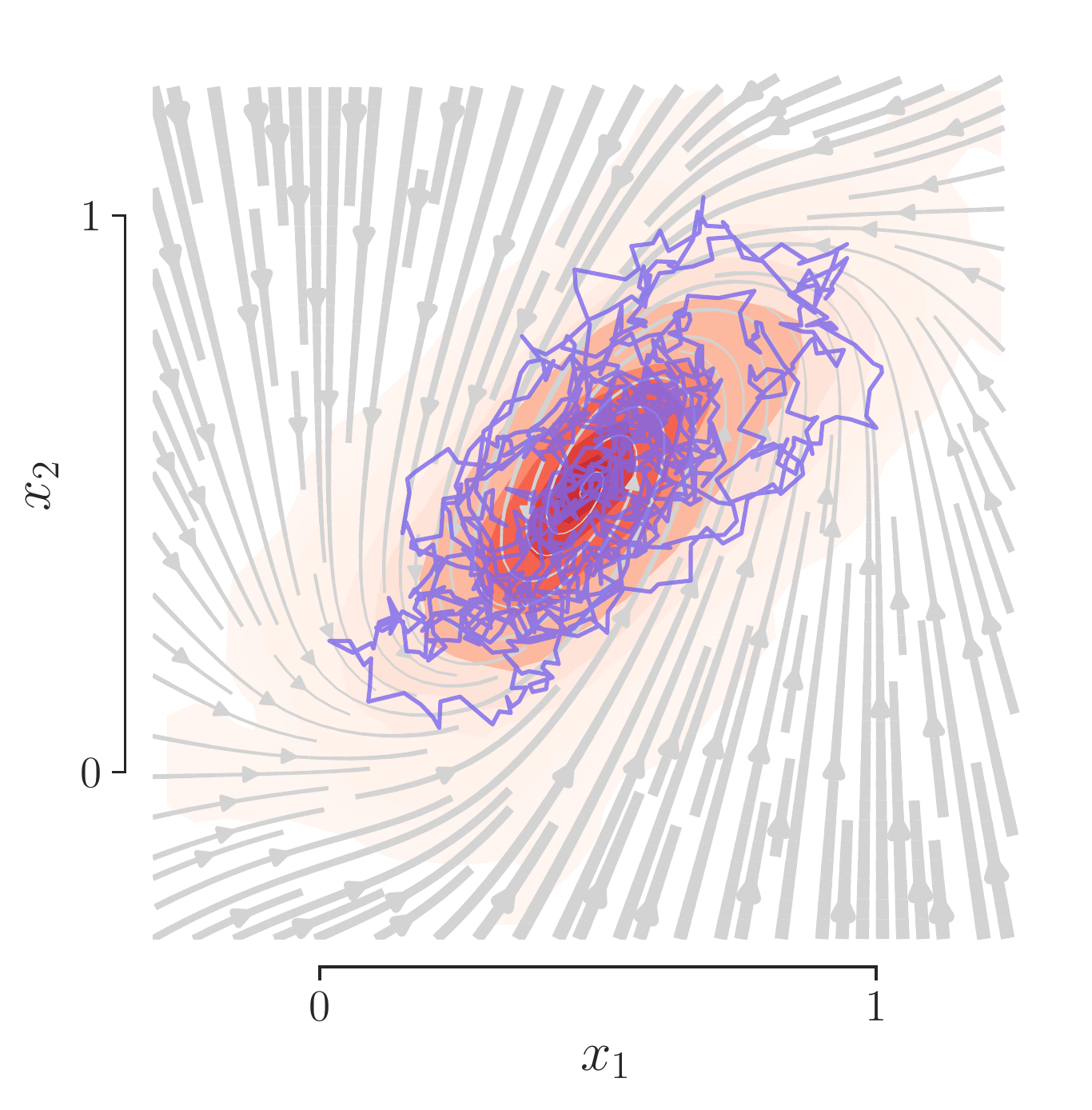}};
    \node at ([yshift=1.6cm,xshift=0.3cm]dyn1.west){\Large{B}};
    \node[anchor=west] (samp1) at (-8,-1.05) {\includegraphics[width=0.495\linewidth]{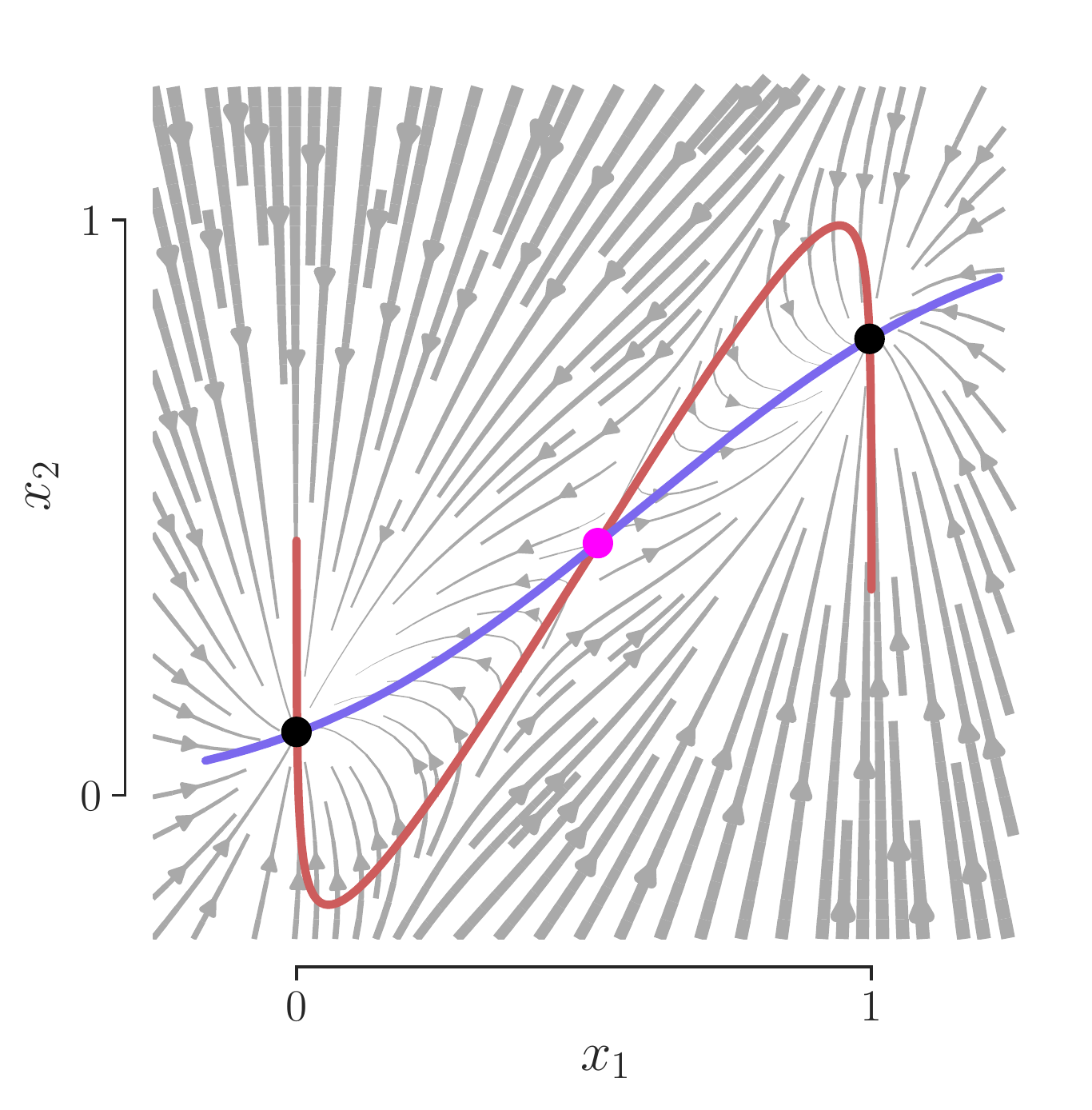}};
    \node[anchor=west] (samp2) at (-4.1,-1.05) {\includegraphics[width=0.495\linewidth]{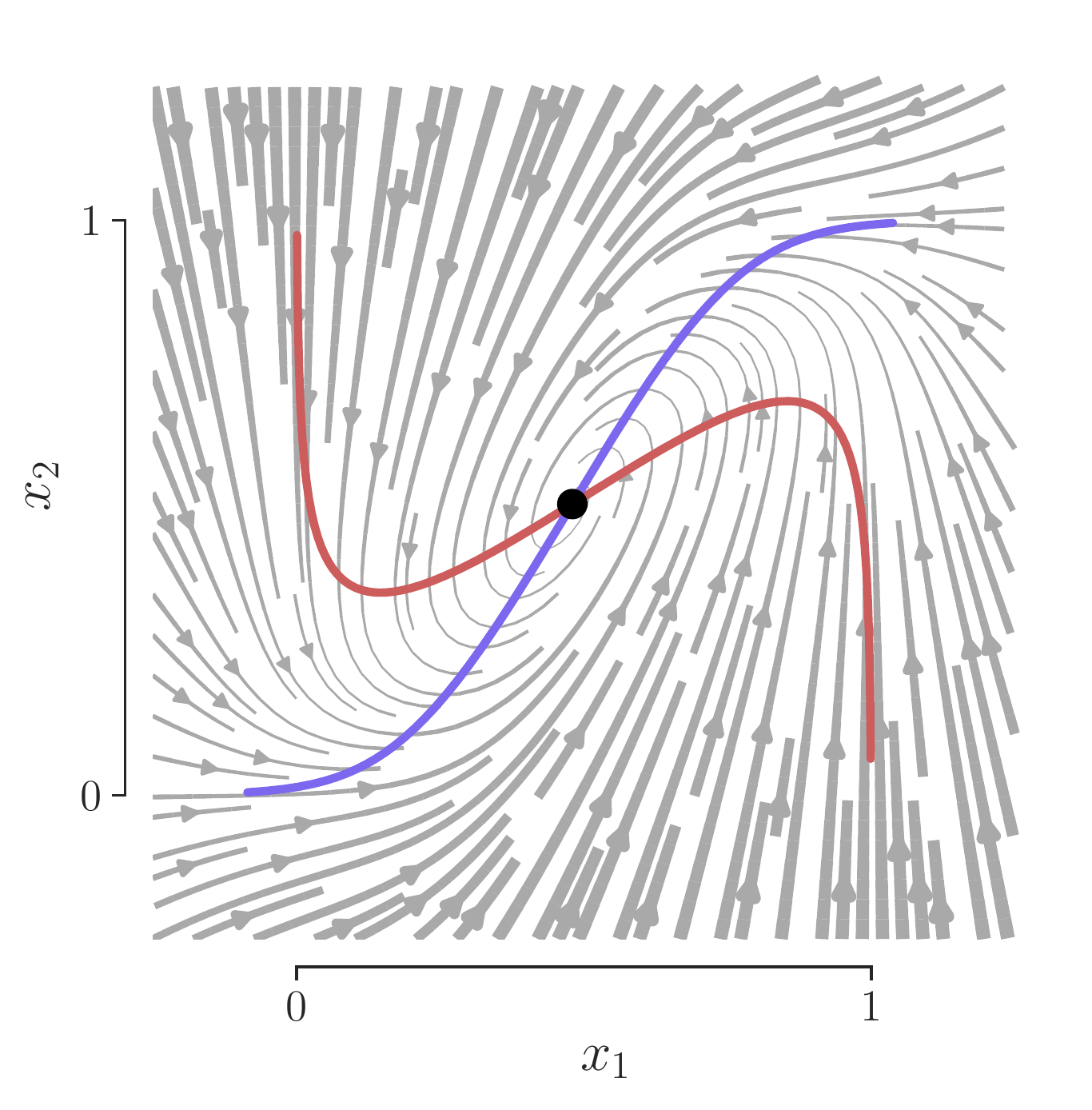}};
    \node at ([yshift=1.6cm,xshift=0.3cm]samp1.west){\Large{C}};
    \node[anchor=west] (est1) at (-8,-4.65) {\includegraphics[width=0.495\linewidth]{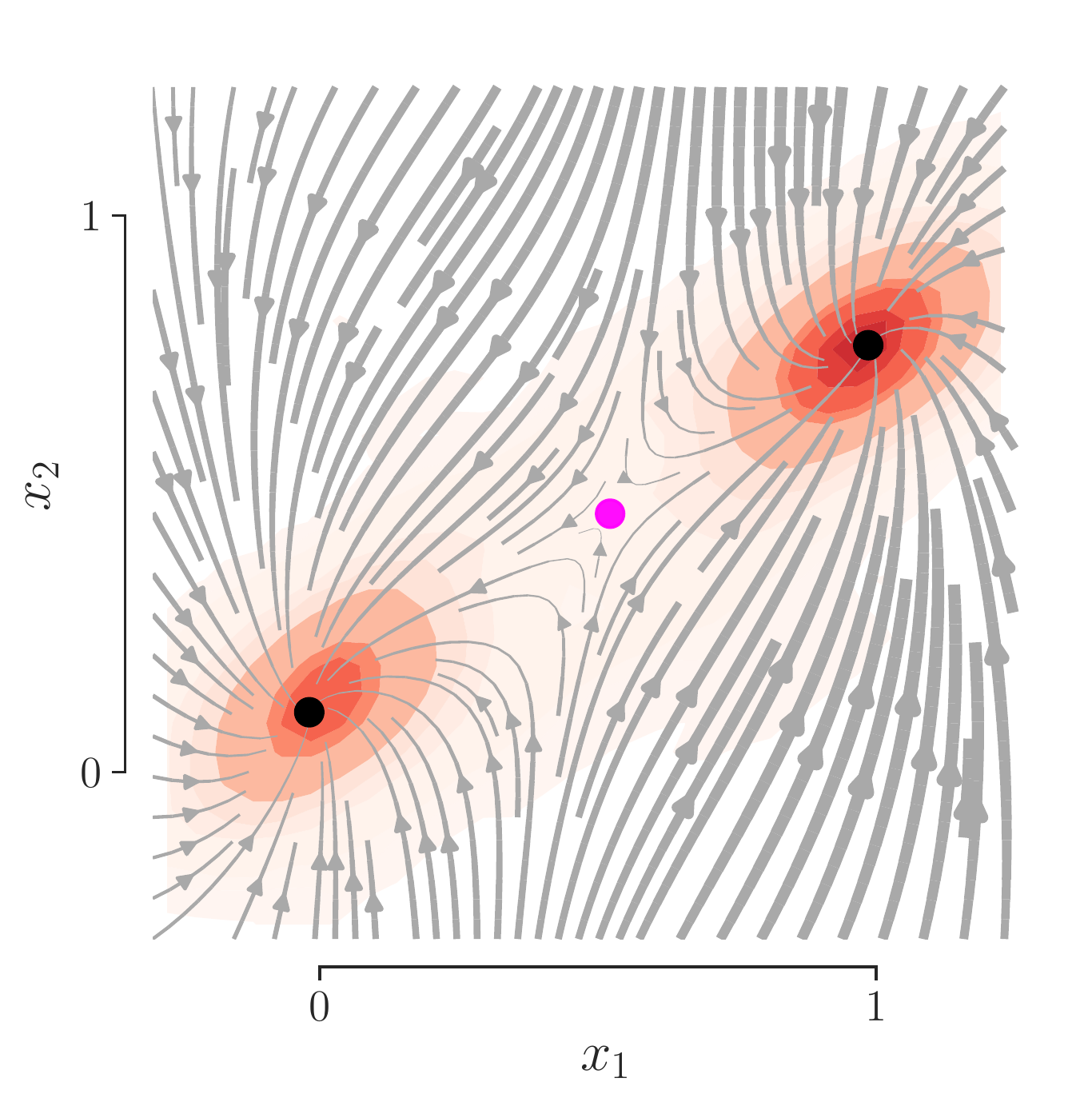}};
    \node[anchor=west] (est2) at (-4.1,-4.65) {\includegraphics[width=0.495\linewidth]{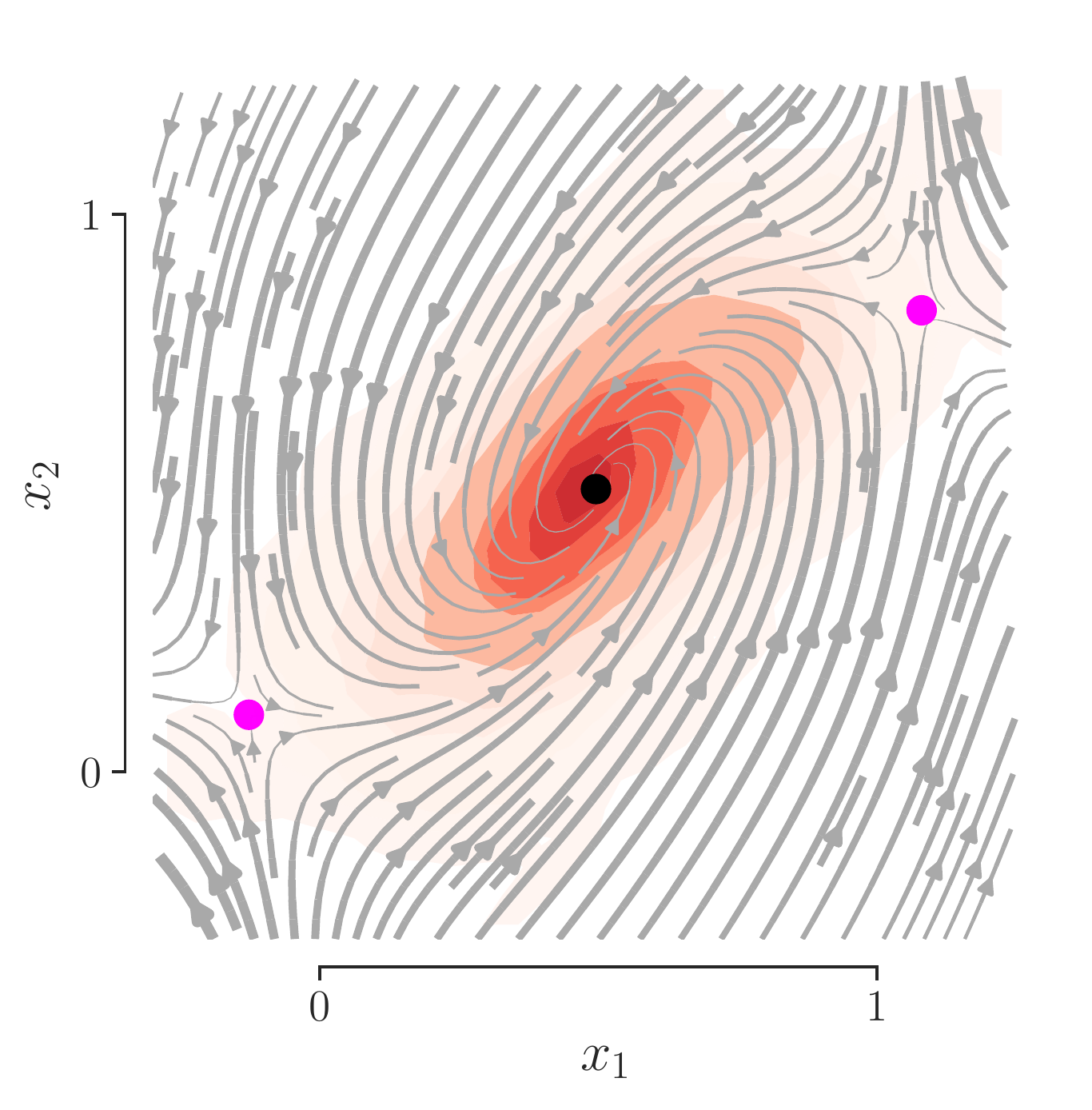}};
    \node at ([yshift=1.6cm,xshift=0.3cm]est1.west){\Large{D}};
    \node[anchor=west] (out2) at (-4.05,5.8) {\includegraphics[width=0.425\linewidth]{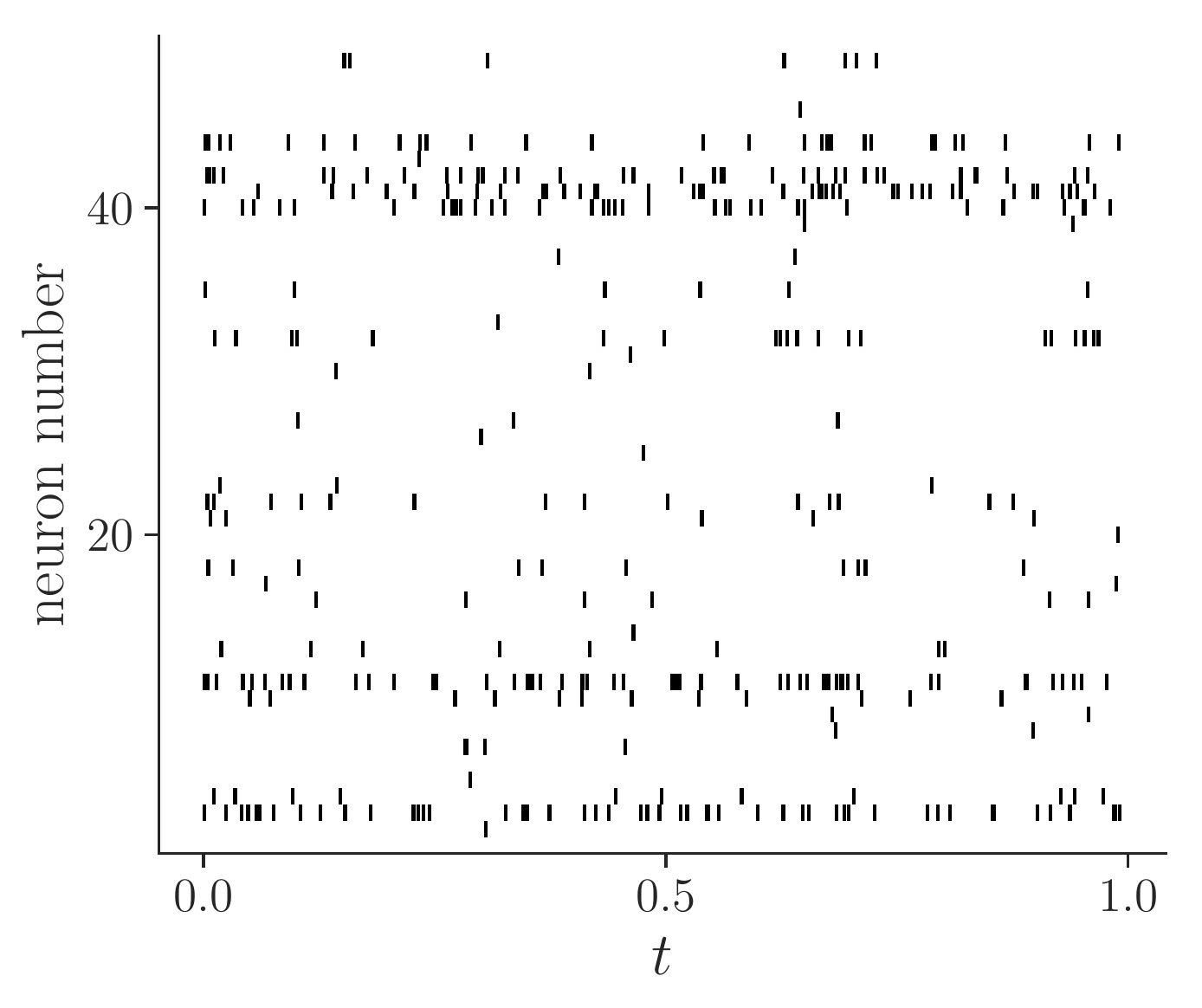}};
    \node[anchor=west] (out1) at (-8.,5.8) {\includegraphics[width=0.425\linewidth]{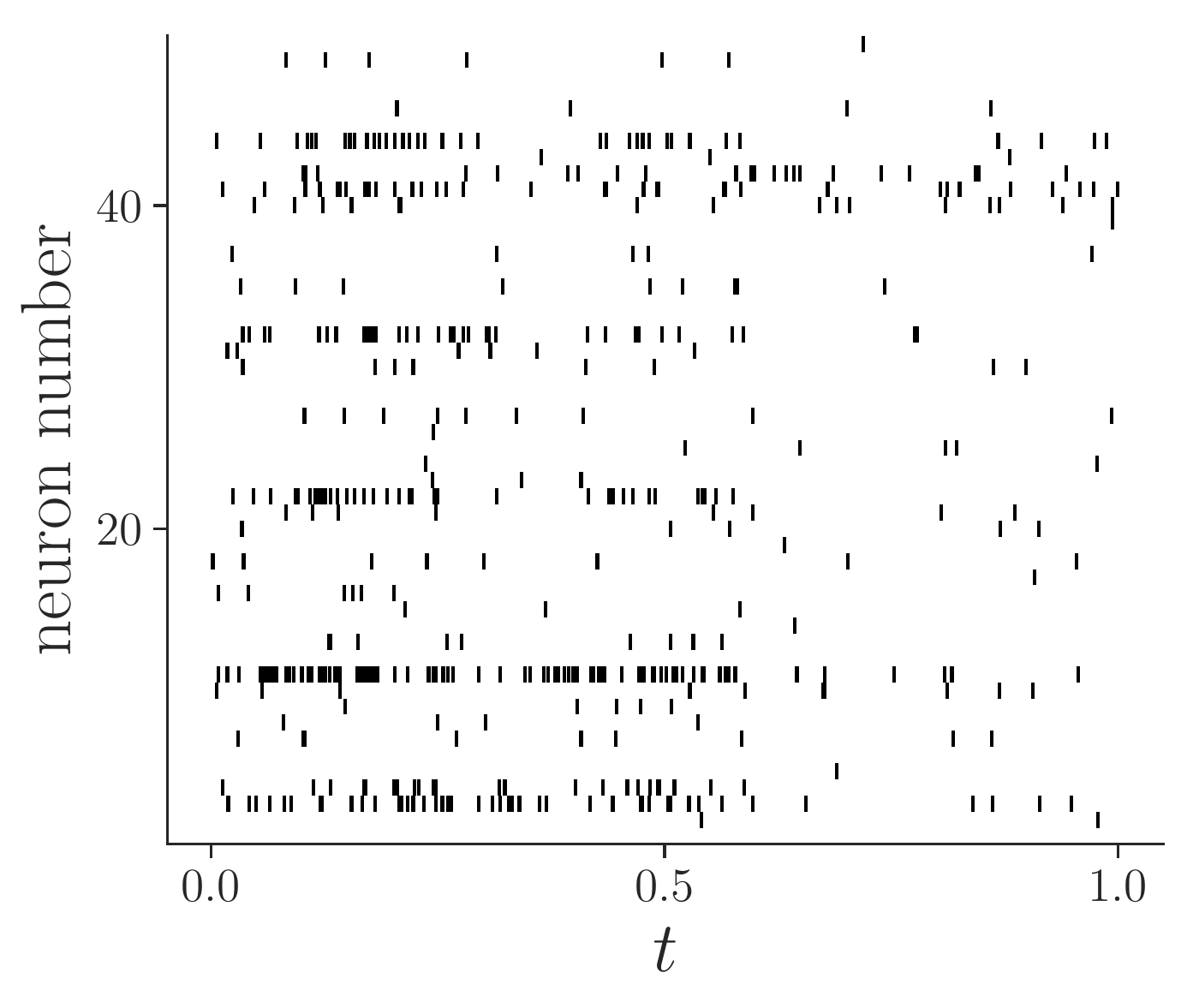}};
    \node at ([yshift=1.2cm,xshift=0.3cm]out1.west){\Large{A}};
\end{tikzpicture}
\caption{Neural population dynamics. Left: simulations with parameter settings $b_1 =1.9  ,b_2 = 0.5, z_1= 3, z_2= 3.9, w_{11}= 10, w_{12}=5 , w_{21}=9 , w_{22}=3 $. Right: simulations with parameter settings $b_1 = 0.4 ,b_2 = 0.6 , z_1= 1.7, z_2=7 , w_{11}= 20, w_{12}=16, w_{21}= 21, w_{22}= 6$. A: Raster plot of the observed spike times for a population of 50 neurons for an example trial. B: Example paths through the two-dimensional latent space on the same trial as A, together with a density plot of latent locations visited across all trials that were used for learning the dynamics, shown in red. C: Streamline plots of the true dynamics together with their fixed points and nullclines for each latent dimension. Stable fixed points are black, unstable ones are magenta. D: Same density plots as in B together with streamline plots of the learnt dynamics and learnt fixed points. The fixed point stability is shown as indicated by the eigenvalues of the learnt Jacobian matrices.}
\label{fig:wilsoncowan}
\end{figure}
This example demonstrates our algorithm under multivariate point-process observations. We model the intensity functions of the $n$th output process as 
$\eta_n(t) = \exp(\sum_{k=1}^K C_{nk} x_k(t) + d_n)$. Conditioned on the intensity function, the $\phi(n)$ observed event-times $\vec t^{(n)}$ are generated by a Poisson process with log likelihood
\begin{equation}
\log p(\vec t^{(n)}| \eta_n) = - \int_\mathcal{T}\eta_n(t)dt  + \sum_{i=1}^{\phi(n)}  \log \eta_n(t_i^{(n)})
\end{equation}
In contrast to the Gaussian observation case, the first term in the log-likelihood above is continuous in $\eta_n(t)$ and the absence of events is also informative towards the underlying intensity of the process.

An interesting application for this setting lies in neural data analysis, where data may be available as a set of spike-times of a population of simultaneously recorded neurons jointly embedded in a circuit involved in performing a computation. In fact, studying neural population activity as a dynamical system has gained increasing traction in the field of neuroscience in recent years \cite{macke+al:2011:nips,shenoy+al:2013:annualreviews,pandarinath+al:2018:naturemethods}, and data analysis methods that can obtain such descriptions are thus of great interest. 

We simulate a two-dimensional latent SDE using the dynamics $f_k(\vec x) = - x_k + \sigma_k( w_{k1} x_1 - w_{k2} x_2  - z_k)$ for $k = 1,2$,
where $\sigma_k(x)=(1 + \exp(-b_k x))^{-1}$. Depending on the choice of parameters $b_k$, $w_{kj}$ and $z_k$ the dynamical system will exhibit different properties. We explore the two regimes where the system either has two stable and one unstable fixed points (Figure \ref{fig:wilsoncowan}C left) or exhibits a single stable spiral (Figure \ref{fig:wilsoncowan}C right).

We simulate data from 50 neurons on 25 trials for each of the two parameter regimes for $b_k$, $w_{kj}$ and $z_k$. Figure \ref{fig:wilsoncowan}A shows example neural spike trains under the two regimes. Figure \ref{fig:wilsoncowan}B illustrates sample paths through the latent space under the different dynamical regimes, together with the density of latent locations visited across all trials. In both settings, we initialise our algorithm with three fixed points and inducing points placed on an evenly spaced $4\times4$ grid in $(-0.25,1.25)$, and hold the parameters relating to the output mapping constant. Figure \ref{fig:wilsoncowan}D shows the estimated flow fields in both settings, together with the location of the fixed points and their stability as indicated by the eigenvalues of the Jacobian matrices. In both settings, our method successfully recovers the main qualitative distinguishing features of the dynamics. In the regime where the dynamics are conditioned on three fixed points but the generative system only contains one, the two additional fixed points will either be associated with higher uncertainty or move to regions where no or little data was observed, as indicated by the superimposed density plots.

\subsection{Multistable chemical reaction dynamics} \label{sec:exp_chem}
This example is based on the dynamical system in \citet{ganapathisubramanian+al:1991:jofchemphys}, which describes nonlinear dynamics of two species of iodione in the iodate-AS(III) system under imperfect mixing by coupled first-order ODEs.  We use these ODEs to describe $\vec f$ and generate data according to (\ref{eq:generative_model}) with high-dimensional Gaussian observations representing spectroscopic measurements, which can approximately be described as a linear mapping from concentrations based on the $\mathrm{I}^-$ and $\mathrm{IO}_3^-$ absorption spectra provided in \citet{kireev+al:2015:ioppub}.
More details on this data-generating process are given in the supplementary material. We simulate data on 20 trials with different initial conditions, collecting 50 unevenly spaced samples from 13 spectroscopy measurements on each trial. 
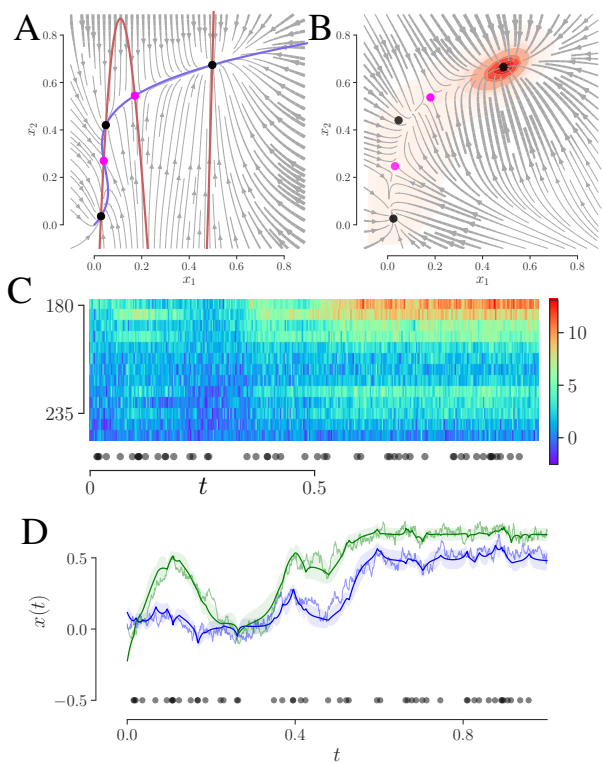
\begin{figure}[h!]
    \centering
    \begin{tikzpicture}
         \node[anchor=west] (dyn) at (-8.6,0) {\includegraphics[width=0.475\linewidth]{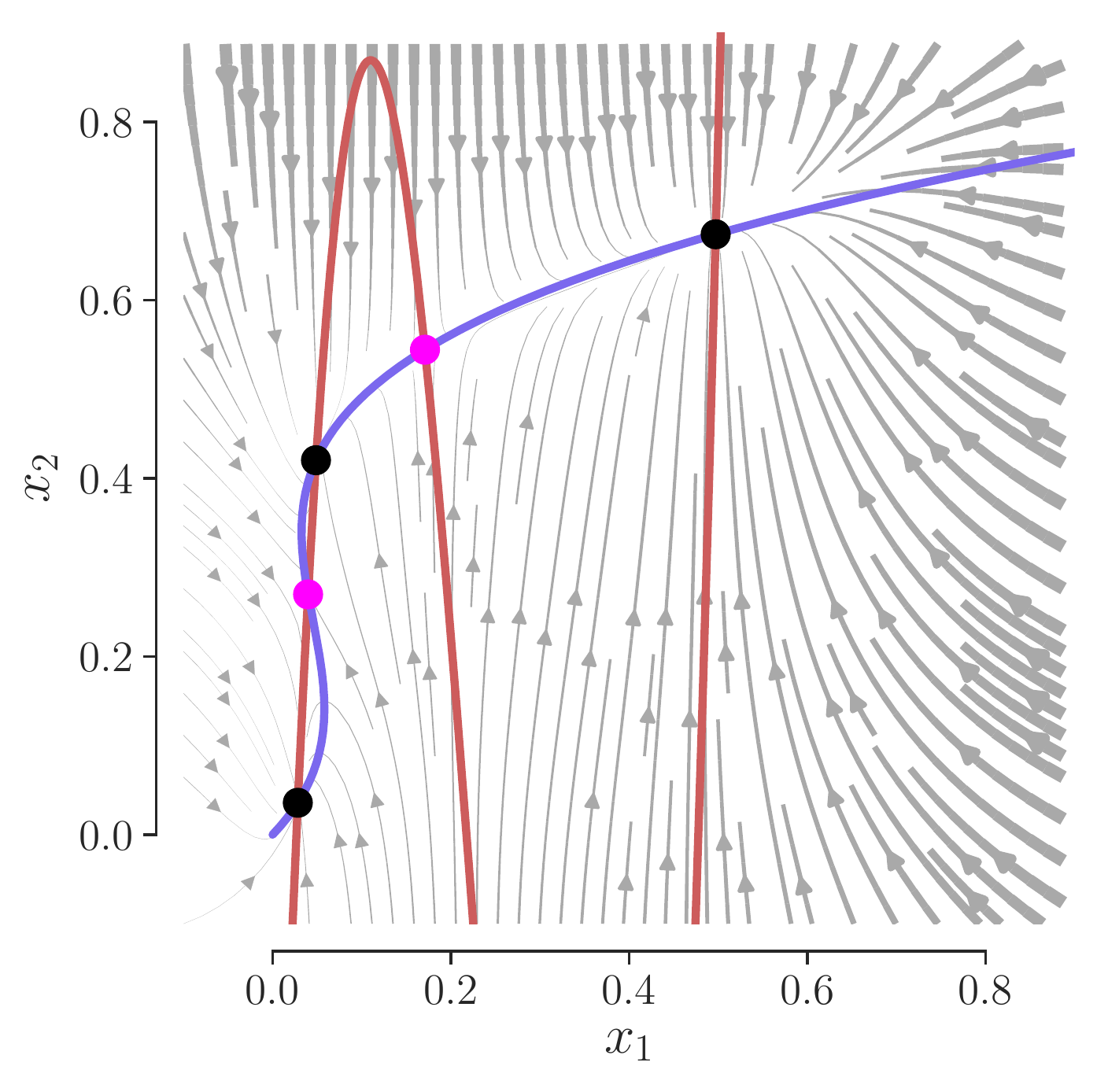}};
         \node at ([yshift=1.6cm,xshift=0.2cm]dyn.west){\Large{A}};
        \node[anchor=west] (est) at (-4.7,0) {\includegraphics[width=0.475\linewidth]{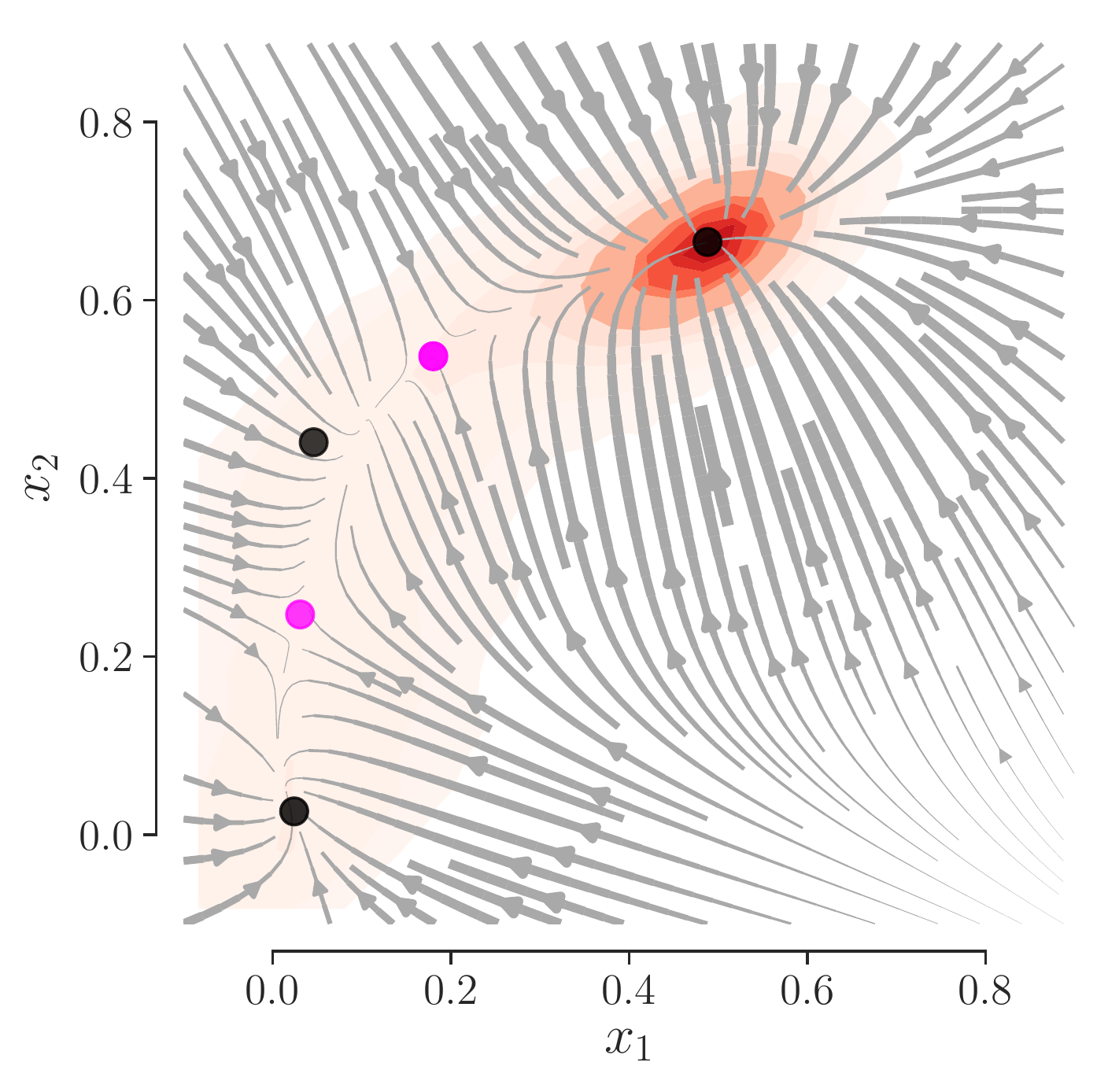}};
         \node at ([yshift=1.6cm,xshift=0.2cm]est.west){\Large{B}};
          \node[anchor=west] (out) at (-8.4,-3.3) {\includegraphics[width=0.9\linewidth]{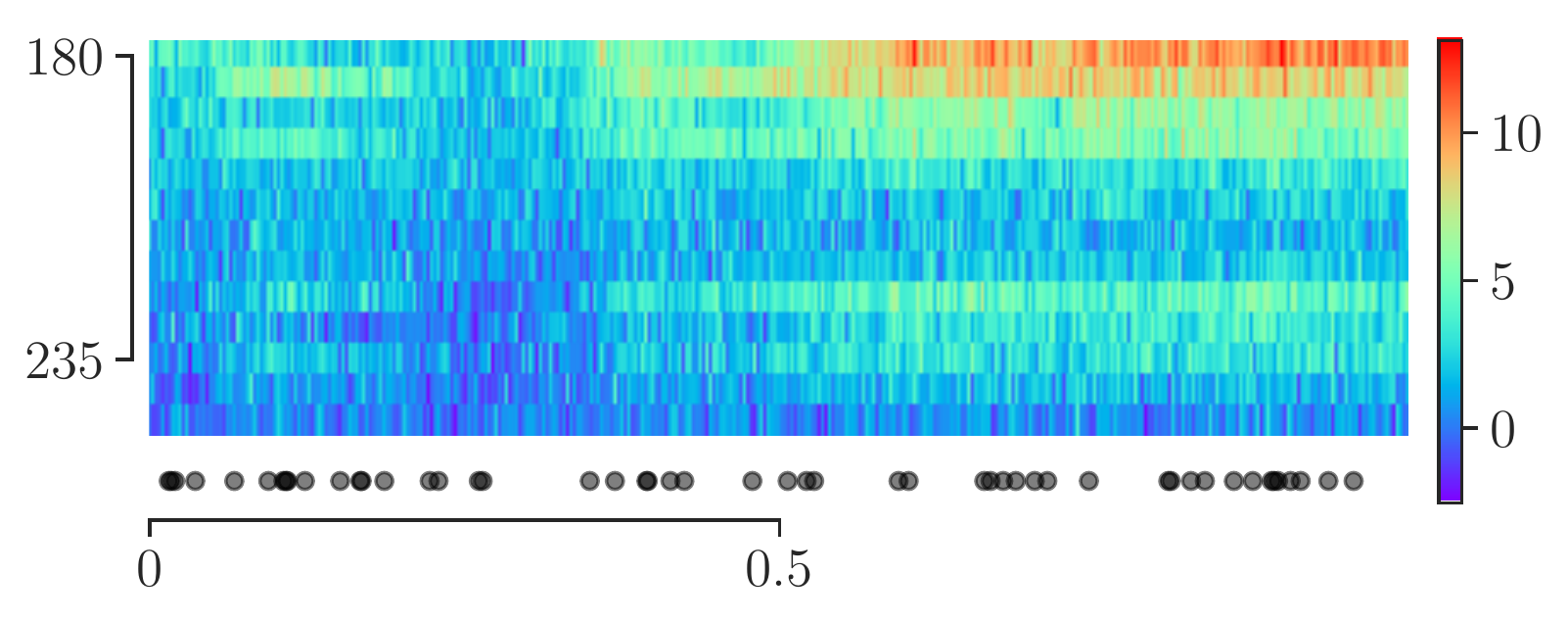}};
         \node at ([yshift=1.4cm,xshift=-0.1cm]out.west){\Large{C}};
         \node at ([yshift=0.4cm,xshift=-1.5cm]out.south){$t$};
         \node[anchor=west] (out) at (-8.6,-6.5) {\includegraphics[width=0.9\linewidth]{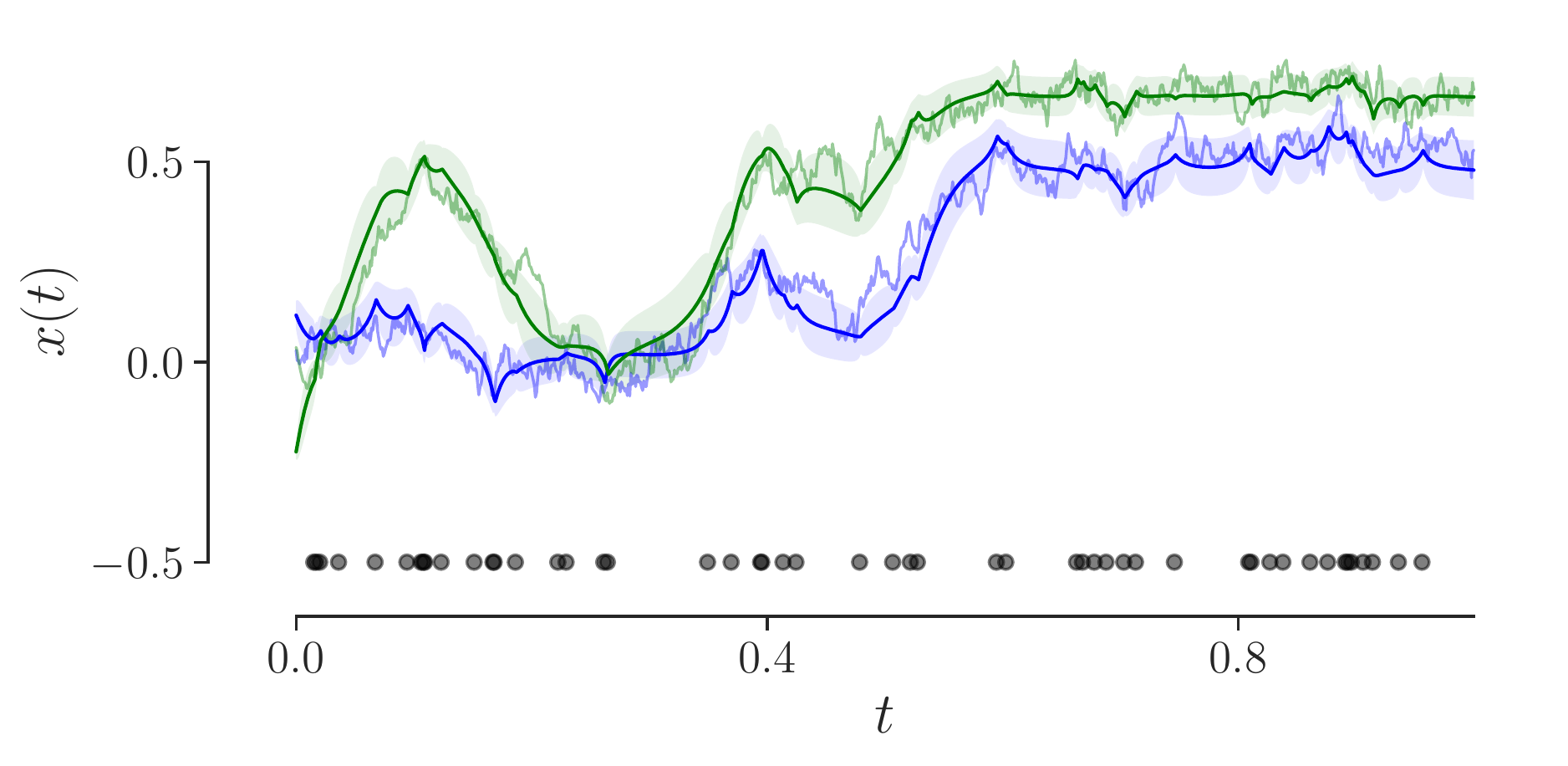}};
         \node at ([yshift=1.4cm,xshift=0.3cm]out.west){\Large{D}};
    \end{tikzpicture}
    \caption{Multistable chemical reaction dynamics. A: Streamline plot of the concentration dynamics for two species of iodine, together with the nullclines and fixed points. Stable fixed points are black, unstable ones are magenta. B: Learnt dynamics and fixed points with stability determined by eigenvalues of learnt Jacobian matrices. Increasing uncertainty in the fixed-point observation is indicated by higher transparency of the dot. The red contour plot illustrates the density of latent path locations across all trials used for training. C: Example spectroscopy measurements (output process) across light wavelengths (nm). D: Example true latent path together with the inferred posterior mean and $\pm$1 standard deviation tubes for each latent dimension on the same trial as C. The black dots indicate the time points at which measurements were taken.}
    \label{fig:chem_react}
\end{figure}

\section{Discussion}

We have introduced a flexible and general variational Bayesian framework for the interpretable modelling of a continuous-time latent dynamical process from intermittent observations.  Using a suitable GP prior, we integrate over a nonparametric description of the system dynamics, conditioned on its fixed points and associated local Jacobian matrices, thus both avoiding the need to assume a specific parametric dynamical form and directly obtaining a meaningful portrait of the dynamical structure. The approach applies to a variety of multivariate observation models, with many updates available in closed form.


The effectiveness of the approach is demonstrated using data simulated from a number of realistic but known nonlinear dynamical systems describing physical, biological and chemical phenomena.  In each case, it was possible to recover a meaningful description of fixed points and nearby dynamics even when data were sparse; and an inferred dynamical model that approximated the true systems well over large regions of the state space.  



A similar prior over dynamics could be adopted within a discrete-time
model such as the GP-SSM, albeit with a less natural interpretation of
the local Jacobians.  However, real-world systems evolve in continuous
time, and in some contexts available observations do not arrive at
discrete sample times.  Retaining a continuous time model means that
the variational posterior over latents can be described by a system of
coupled ODEs.  While the solution of these may incur a discretisation
error, this is a numerical issue related to the choice of ODE solver,
rather than the assumption of a discretised model.  Indeed, the ODE
solution can exploit an adaptive step size in a way that would be
impractical within a discrete-time model.


Our work also differs from other GP-based approaches to time series modelling, where each dimension of the process $x_k(t)$ is modelled via an independent GP \cite{damianou+al:2011:nips, duncker+al:2018:nips}. In this case, the prior on $\vec x(t)$ evaluated at any finite set of points can be described by a multivariate Gaussian distribution, which greatly simplifies the inference. However, this cannot capture correlations across the dimensions of the latent process and thus comes at a loss of the descriptive power.

The variational inference approach for SDEs from \citet{archambeau+al:2007:jmlr, archambeau+al:2008:nips} relied on a Gaussian observation model and known dynamics \cite{archambeau+al:2007:jmlr}, or a known parameterisation of the dynamics \cite{archambeau+al:2008:nips}, both of which are restrictive. Here, we have extended the inference approach to handle a wider class of observation models, as well as a nonparametric GP description of the dynamics. \citet{batz+al:2018:physreview} also use a GP to model the drift function of an SDE. However, they consider the setting where dense or sparse observations of the SDE path are directly available, while we treat the entire SDE as latent. Furthermore, the interpretable nonparametric representation of the SDE dynamics in terms of their fixed points and local Jacobian matrices is novel.

While we have demonstrated our algorithm in the setting of unevenly sampled multivariate Gaussian and multivariate point process observations, the inference approach extends readily to other stochastic processes typically considered challenging to model, such as marked point processes. We therefore expect this approach to have diverse applications, ranging from neuroscience to chemistry and finance. 

\section*{Acknowledgements}
This work was funded by the Simons Foundation (SCGB 323228, 543039; MS) and the
Gatsby Charitable Foundation.

\bibliography{refs}
\bibliographystyle{icml2019}






\appendix
\newpage
\appendix
\section{Variational Lower Bound}
We derive a variational lower bound to the marginal log-likelihood of our model using Jensen's inequality
\begin{align*}
    & \log p(\vec y|\vec \theta) = \log \int d\vec x d \vec f d\vec u \; p(\vec y| \vec x) p(\vec x | \vec f) p(\vec f|\vec u,\vec \theta) p(\vec u| \vec \theta)\\
    & \ge \int  d\vec x d \vec f d\vec u q(\vec x,\vec f, \vec u) \log \frac{p(\vec y| \vec x) p(\vec x | \vec f) p(\vec f|\vec u,\vec \theta) p(\vec u| \vec \theta)}{q(\vec x,\vec f, \vec u)}\\
    & \qquad \overset{def}{=} \mathcal{F^*}
\end{align*}
where $p(\vec f|\vec u,\vec \theta) p(\vec u| \vec \theta) = \prod_{k} p(f_k | \vec u_k, \vec \theta) p(\vec u_k|\vec \theta)$. Choosing a factorised variational distribution of the form
\[
q(\vec x,\vec f, \vec u) =  q_x(\vec x) \prod_{k} p(f_k | \vec u_k, \vec \theta) q_{u}(\vec u_k)
\]
we can rewrite the bound as
\begin{align*}
    \mathcal{F}^* &= \int  d\vec x d \vec f d\vec u q(\vec x,\vec f, \vec u) \log \frac{p(\vec y| \vec x) p(\vec x | \vec f) \prod_{k} p(\vec u_k|\vec \theta) }{q_x(\vec x) \prod_{k} q_{u}(\vec u_k)} \\
    & = \left\langle \log p(\vec y| \vec x) \right\rangle_{q_x} - \left\langle \KL{q_x(\vec x)}{p(\vec x| \vec f)} \right\rangle_{q_f}\\
    & \qquad - \sum_k \KL{q_u(\vec u_k)}{p(\vec u_k)}
\end{align*}
where 
\[
q_f(\vec f) = \prod_k \int d \vec u_k  p(f_k | \vec u_k, \vec \theta) q_{u}(\vec u_k)
\]
and $q_x(\vec x)$ is described by (\ref{eq:q_sde}),(\ref{eq:q_marginal}) and (\ref{eq:q_odes}).
We can derive the Kullback-Leibler divergence between the distributions over SDE paths $q_x(\vec x)$ and $p(\vec x|\vec f)$ by discretising time in steps of $\Delta t$. The discretised paths have Markovian structure with
\begin{align*}
p(\vec x_{t+1}|\vec x_t, \vec f) & = \mathcal{N}(\vec x_{t+1} | \vec x_t + \vec f(\vec x_t) \Delta t, 
	\vec \Sigma \Delta t)\\
q_x(\vec x_{t+1}|\vec x_t) &= \mathcal{N}(\vec x_{t+1} |  \vec x_t +  \vec f_q(\vec x_t) \Delta t, \vec \Sigma \Delta t)
\end{align*}
We can hence write
\begin{align*}
	 & \KL{q_x(\vec x)}{p(\vec x)} \\
	 & = \sum_{t=1}^{T-1} \int d\vec x_t q(\vec x_t) \int d \vec x_{t+1} q(\vec x_{t+1}|\vec x_t) \log \frac{q(\vec x_{t+1}|\vec x_t) }{p(\vec x_{t+1}|\vec x_t)}\\
	& =  \half \sum_{t=1}^{T-1} \Delta t \left \langle (\vec f - \vec f_q)\tr \vec \Sigma\inv (\vec f - \vec f_q)  \right \rangle_{q_X}
\end{align*}
Taking the limit as $\Delta t \to 0$, we obtain
\[
 \KL{q_x(\vec x)}{p(\vec x)} =  \half \int_{\mathcal{T}} dt \left \langle (\vec f - \vec f_q)\tr \vec \Sigma\inv (\vec f - \vec f_q)  \right \rangle_{q_x}
\]

\section{Inference Details}
\subsection{Lagrangian}
The full Lagrangian, after applying integration by parts to the constraints in (\ref{eq:Lagrangian_constraints}), has the form
\begin{align*}
    \mathcal{L} & =  \mathcal{F}^*  - \mathcal{C}_1 - \mathcal{C}_2 \\
    \mathcal{C}_1 & = \int_{\mathcal{T}}dt\; \left(\trace{\vec \Psi(\vec A\vec S_x + \vec S_x \vec A\tr -I) - \frac{d\vec \Psi}{dt} \vec S_x}\right) \\
    & \qquad + \trace{\vec \Psi(T)\vec S_x(T)} - \trace{\vec \Psi(0)\vec S_x(0)}  \\
    \mathcal{C}_2 & = \int_{\mathcal{T}}dt\; \left( \vec \lambda\tr (A \vec m_x -\vec b) - \frac{d \vec \lambda}{d t} \tr \vec m_x \right) \\
    & \qquad + \vec\lambda(T)\tr \vec m_x(T) -  \vec\lambda(0)\tr \vec m_x(0)
\end{align*}
For the variational free energy term $\mathcal{F}^*$, we have from before
\[
\mathcal{F} = \sum_i \left\langle \log p(\vec y_i | \vec x_i)\right \rangle_{q_x} - \KL{q_x(\vec x)}{p(\vec x)}
\]
and 
\[
\mathcal{F}^* = \left\langle \mathcal{F} \right\rangle_{q_{f}} - \sum_{k=1}^K \KL{q_u(\vec u_k)}{p(\vec u_k| \vec \theta)}
\]
The Kullback-Leibler divergences can be evaluated as
\begin{align*}
	\KL{q(\vec u_k)}{p(\vec u_k | \vec \theta)} & = \half \Big( \trace{{\vec \Omega_u} \inv \vec S_u^k} - M + \log \frac{ |\vec \Omega_u|} { |\vec S_u^k|}\\
	& + (\vec \mu_u^k - \vec m_u^k)\tr {\vec \Omega_u}\inv (\vec \mu_u^k - \vec m_u^k)  \Big)
\end{align*}
with
\begin{align*}
	\vec \Omega_u & = \vec K_{zz} - \tilde{\vec K}_{zs}  {\tilde{\vec K}_{ss}}\inv  \tilde{\vec K}_{sz}  \\
	\vec \mu_u^k & =  \tilde{\vec K}_{zs} {\tilde{\vec K}_{ss}}\inv \vec v_k^\theta
\end{align*}
and
\begin{align*}
	& \left \langle \KL{q_x(\vec x)}{p(\vec x)} \right\rangle_{q_f}  = \half \int_0^T dt \left \langle \right (\vec f - \vec f_	q)\tr (\vec f - \vec f_	q) \rangle_{q_x q_f} 
\end{align*}
For later convenience, we denote this term as
\[
\left \langle \KL{q_x(\vec x)}{p(\vec x)} \right\rangle_{q_f} = \mathcal{E}(\vec m_x, \vec S_x)
\]
Using the identity
\[
\langle \langle \vec f \rangle_{q_f} \left( \vec x - \vec m_x \right)\tr \rangle_{q_x} = \left \langle \frac{\partial \langle \vec f \rangle_{q_f} }{\partial \vec x} \right\rangle_{q_x} \vec S_x
\]
the integrand can be evaluated as
\begin{align*}
& \left \langle \right (\vec f - \vec f_	q)\tr (\vec f - \vec f_	q) \rangle_{q_x q_f} \\
& = \langle\vec f\tr \vec f\rangle _{q_x q_f} +2 \trace{\vec A\tr \left\langle \frac{ \partial \vec f}{\partial \vec x} \right\rangle_{q_x q_f} \vec S(t) } \\
& \qquad + \trace{\vec A \tr \vec A \left(\vec S_x + \vec m_x\vec m_x\tr \right)} + 2 \; \vec m_x\tr \vec A \tr \langle \vec f \rangle_{q_x q_f}  \\
& \qquad  + \vec b \tr \vec b -2 \vec b \tr \langle \vec f\rangle - 2 \vec b\tr\vec A \vec m_x
\end{align*}

For the expected log-likelihood terms, in general, there will be terms that are continuous in $\vec x$, and terms that depend only on evaluations of $\vec x$ at specific locations $t_i$, which we will denote by $\ell^{cont}$ and $\ell^{jump}$, respectively. Such that we can write
\[
\langle \log p(\vec y|\vec x)\rangle_{q_x} = \ell^{cont}(\vec m_x, \vec S_x)+ \ell^{jump}(\vec m_x, \vec S_x)
\]
 Thus, the variational free energy can be expressed as
\begin{align*}
\mathcal{F}^* & = \ell^{cont}(\vec m_x, \vec S_x)+ \ell^{jump}(\vec m_x, \vec S_x) -  \mathcal{E}(\vec m_x, \vec S_x) \\
& \qquad - \sum_{k=1}^K \KL{q_u(\vec u_k)}{p(\vec u_k| \vec \theta)}
\end{align*}

\subsubsection{Example: Gaussian likelihood}
In the case of a Gaussian likelihood, there is no continuous term in the likelihood such that
\begin{align*}
    \ell^{cont} & = 0 \\ 
    \ell^{jump} & =  \sum_i  \int_{\mathcal{T}_0}^{\mathcal{T}_{end}} dt \delta(t - t_i) \big(\vec m_x(t)\tr \vec C\tr \Gamma\inv (\vec y_t - \vec d) \\
    & -\half \trace{\vec C\tr \Gamma\inv \vec C \sum_i \left(\vec S_x(t) +\vec m_x(t)\vec m_x(t)\tr \right)} \big)
\end{align*}

\subsubsection{Example: Multivariate Poisson Process likelihood}
In the case of a multivariate Poisson Process, with $g(\cdot) = \exp(\cdot)$ and observed event times $t_1^{(n)}, \dots t_{\phi(n)}^{(n)} $for the $n$th output dimension
\begin{align*}
    \ell^{cont} & = - \sum_n \int_{\mathcal{T}_0}^{\mathcal{T}_{end}} \exp\left( \vec c_n \tr \vec m_x + \half \vec c_n \tr \vec S_x \vec c_n \right)dt \\ 
    \ell^{jump} & =  \sum_{n=1}^N \sum_{i=1}^{\phi(n)} \int_{\mathcal{T}_0}^{\mathcal{T}_{end}} \left( \vec c_n \tr \vec m_x(t) + d_n\right)\delta(t - t_i^{(n)})dt
\end{align*}

\subsection{Symmetric variations in $\vec S_x$}
To arrive at the fixed point equations given in the main paper, we need to take variational derivatives of the Lagrangian with respect to $\vec m_x$ and $\vec S_x$. In contrast to \cite{archambeau+al:2007:jmlr}, we take the symmetric variations in $\vec S_x$ into account. Also note that the Lagrange multiplier $\vec \Psi$ is symmetric. We can hence write
\begin{align*}
 \frac{\partial \mathcal{C}_1}{\partial \vec S_x} & = \left(\vec \Psi \vec A + \vec A\tr \vec \Psi  - \frac{d \vec\Psi}{dt}\right) \odot \tilde{\mathbb{P}}
\end{align*}
where $\odot$ denotes the elementwise Hadamard product and $\tilde{\mathbb{P}}_{ij} = 2$ for $i \ne j$ and $1$ otherwise. Differentiating the entire Lagrangian with respect to the symmetric matrix $\vec S_x$ and setting to zero we get
\begin{align*}
0 =  \frac{\partial \mathcal{F}^*}{\partial \vec S_x} \odot \mathbb{P}  - \vec \Psi \vec A - \vec A\tr \vec \Psi  + \frac{d \vec\Psi}{dt}
\end{align*}
matching the equation given in the main text with  $\mathbb{P}_{ij} = \half$ if $i\ne j$ and $1$ otherwise. Note that the derivatives of the free energy with respect to $\vec S_x$ will also need to take into account the symmetry of the covariance matrix. The derivations for (\ref{eq:lambda_ode})-(\ref{eq:b_update}) follow those of \citet{archambeau+al:2007:jmlr}.

\subsection{Expected values of dynamics}
The inference algorithm requires evaluating several expectations with respect to $q_x$ and $q_f$. Let $\vec U = \begin{bmatrix}
\vec u_1&  \dots &\vec u_K 
\end{bmatrix}$ and $\langle \vec U \rangle_{q_u} = \vec M_u$, such that we can define  $(M + L + LK) \times K$ matrices stacking all inducing points, zero function values, and Jacobians as
\begin{align*}
\vec U_{u,f_s,J} & = 
\begin{bmatrix}
    \vec U_u\\
    \vec 0 \\
    \vec J_s^{(1)}\\
    \vdots\\
    \vec J_s^{(L)}
    \end{bmatrix},
&&\langle \vec U_{u,f_s,J} \rangle_{q_u} = \vec M_{u,f_s,J} = 
\begin{bmatrix}
    \vec M_u\\
    \vec 0 \\
    \vec J_s^{(1)}\\
    \vdots\\
    \vec J_s^{(L)}
    \end{bmatrix}
\end{align*}
The required expectations can then be evaluated as 
\begin{align*}
 \left\langle \vec f(\vec x) \right\rangle_{q_x q_f}\tr &= \left\langle \vec a_z^\theta (\vec x)\right\rangle_{q_x}  \vec M_{u,f_s,J}
\end{align*}
\begin{align*}
\left\langle \frac{\partial \vec f(\vec x)}{\partial \vec x} \right\rangle_{q_x q_f}\tr &= \left\langle \nabla_x \vec a_z^\theta (\vec x)\right\rangle_{q_x}  \vec M_{u,f_s,J}
\end{align*}
\begin{align*}
 & \left\langle \vec f(\vec x)\tr \vec f(\vec x) \right\rangle_{q_x q_f}  = \sum_k\left\langle f_k^2(\vec x) \right\rangle_{q_x q_f} =   \kappa(\vec x, \vec x') \\
 & + \trace{ \left( \left\langle \vec U_{u,f_s,J} \vec U_{u,f_s,J}\tr \right\rangle_{q_u} - {\mat K_{zz}^\theta}\right) \langle \vec a_z^\theta (\vec x)\tr \vec a_z^\theta (\vec x) \rangle_{q_x}}
\end{align*}
The above expressions still involve computing expectations of covariance functions and their derivatives, which can be computed analytically for choices such as the exponentiated quadratic covariance function.

\subsection{Inference algorithm}
The full inference algorithm involves solving a set of ODEs forward and backward in time, which we do using the forward Euler method. We provide the full approach in Algorithm \ref{alg:smoothing}, where the subscript $r$ denotes the evaluation of the functions at the $r$th point of the time grid between $\mathcal{T}_0$ and $\mathcal{T}_{end}$ taking steps of size $\Delta t$. Note that the derivatives of the terms in $\ell^{jump}$ will need to be discretized appropriately as well. Using the same time-grid as was used for solving the ODEs, the delta-functions will contribute a factor of $\frac{1}{\Delta t}$, such that the $\Delta t$ terms cancel in the update written in Algorithm \ref{alg:smoothing}.

\begin{algorithm*}[tb]
  \caption{Inference algorithm}
  \label{alg:smoothing}
\begin{algorithmic}
  \STATE {\bfseries Input:} data $\{y_i, t_i\}_{i=1}^T$, $\vec m_{x,0}$, $\vec S_{x,0}$, $q_f(\vec f)$, $\Delta t$, $\mathcal{T}_0$, $ \mathcal{T}_{end}$
  \STATE Initialize $\vec A(t), \vec b(t)$
  \STATE $R = \frac{\mathcal{T}_0 -  \mathcal{T}_{end}}{\Delta t}$
  \REPEAT
  \FOR{$r=0$ {\bfseries to} $R-1$}
  \STATE $\vec m_{x,{r+1}} \leftarrow \vec m_{x,r} - \Delta t\left(\vec A_{r}  \vec m_{x,r} - \vec b_{r}\right)$
  \STATE $\vec S_{x,{r+1}} \leftarrow \vec S_{x,r} - \Delta t\left(\vec A_{r}  \vec S_{x,r} + \vec S_{x,r} {\vec A_{r}}\tr - I\right)$
  \ENDFOR
  \FOR{$r=R$ {\bfseries to} $1$}
  \STATE $\vec \lambda_{r-1} \leftarrow \vec \lambda_{r} - \Delta t\left( {\vec A_{r}}\tr \vec \lambda_r + \left(\frac{\partial \ell^{cont}}{\partial \vec m_x} - \frac{\partial \mathcal{E}}{\partial \vec m_x}\right)|_{t = r\Delta t}\right) - \Delta t\frac{\partial \ell^{jump}}{\partial \vec m_x}\Big|_{t=(r-1)\Delta t}$
  \STATE $\vec \Psi_{r-1} \leftarrow \vec \Psi_{r} - \Delta t\left( {\vec A_{r}}\tr \vec \Psi_{r} + \vec \Psi_{r} \vec A_{r} +  \mathbb{P} \odot \left(\frac{\partial \ell^{cont}}{\partial \vec S_x} -  \frac{\partial \mathcal{E}}{\partial \vec S_x}\right)|_{t = r\Delta t} \right) -  \Delta t\mathbb{P} \odot \frac{\partial \ell^{jump}}{\partial \vec S_x}\Big|_{t=(r-1)\Delta t} $  
  \ENDFOR
  \STATE $\vec A= \left\langle \frac{\partial \vec f}{\partial \vec x} \right\rangle_{q_x q_f}  + 2 \vec\Psi $
   \STATE $\vec b = \left\langle \vec f(\vec x) \right\rangle_{q_x q_f} + \vec A \vec m_{x} - \vec\lambda  $ 
  \UNTIL{convergence in $\mathcal{F}^*$}
  \STATE {\bfseries return:} $\{\vec A_r, \vec b_r, \vec \lambda_r, \vec \Psi_r, \vec m_{x,r}, \vec S_{x,r}\}_{r=1}^R$
\end{algorithmic}
\end{algorithm*}

\section{Learning Details}
\subsection{Conditioned Sparse Gaussian Process dynamics}
The only term in the variational free energy that depends on the parameters in $\vec f$ are the KL-divergence between the continuous-time processes and the KL-divergence relating to the inducing points for $\vec f$.
\subsubsection{Inducing point covariances}
Collecting the terms that contain $\vec S_u^k$ we have
\begin{align*}
   \frac{\partial}{\partial \vec S_u^k} \KL{q(\vec u_k)}{p(\vec u_k | \vec \theta)}  & = \half {\vec \Omega_u} \inv - \half {\vec S_u^k}\inv
\end{align*}
\begin{align*}
   \frac{\partial \mathcal{E}}{\partial \vec S_u^k} & = \half\int_\mathcal{T} dt \frac{\partial}{\partial \vec S_u^k} \trace{ \begin{bmatrix}
   \vec S_u^k & 0 \\
   0 & 0
   \end{bmatrix}  \langle \vec a_z^\theta (\vec x)\tr \vec a_z^\theta (\vec x) \rangle_{q_x} }\\
   & = \half \int_\mathcal{T} dt  [ \langle \vec a_z^\theta (\vec x)\tr \vec a_z^\theta (\vec x) \rangle_{q_x}]_{:M,:M}
\end{align*}
where the last line selects the first $M\times M$ block from $\langle \vec a_z^\theta (\vec x)\tr \vec a_z^\theta (\vec x) \rangle_{q_x}$. We hence obtain the closed form update 
\[
\vec S_u^k = \left( {\vec \Omega_u} \inv +  \int_\mathcal{T} dt  [ \langle \vec a_z^\theta (\vec x)\tr \vec a_z^\theta (\vec x) \rangle_{q_x}]_{:M,:M} \right) \inv 
\]

\subsubsection{Inducing points and Jacobians}
To find the update efficiently, let $\vec J_k = [ \vec J^{(1)}_{k,:}, \dots, \vec J^{(L)}_{k,:}]\tr$ so that we can write
\[
\vec \mu_u^k  =  \tilde{\vec K}_{zs} {\tilde{\vec K}_{ss}}\inv \vec v_k^\theta = \tilde{\vec K}_{zs} {\tilde{\vec K}_{ss}}\inv \begin{bmatrix}
\vec 0\\
\vec J_k
\end{bmatrix} = \vec G \vec J_k
\]
We can rewrite the quadratic terms in the Kullback-Leibler divergences of the inducing points as
\begin{align*}
&\sum_k (\vec \mu_u^k - \vec m_u^k)\tr {\vec \Omega_u}\inv (\vec \mu_u^k - \vec m_u^k)  \\
& \qquad =  \sum_k \begin{bmatrix}
\vec {m}_u^k\\
\vec J_k
\end{bmatrix}\tr
 \begin{bmatrix}
{\vec \Omega_u}\inv & - {\vec \Omega_u}\inv \vec G\\
- \vec G\tr {\vec \Omega_u}\inv & \vec G \tr {\vec \Omega_u}\inv \vec G
\end{bmatrix}
 \begin{bmatrix}
\vec {m}_u^k\\
\vec J_k
\end{bmatrix}\\
& \qquad = \trace{\vec  M_{u,J}\tr \tilde{\vec \Omega} \vec M_{u,J}}
\end{align*}
with $\vec M_{u,J} = \begin{bmatrix}
\vec m_u^1&  \dots &\vec m_u^K \\
\vec J_1 & \dots & \vec J_K
\end{bmatrix}$ and derivative
\begin{align*}
	\frac{\partial}{\partial \vec M_{u,J}} \sum_k \KL{q(\vec u_k)}{p(\vec u_k | \vec \theta)} & =  \tilde{\vec \Omega} \vec M_{u,J}
\end{align*} 
\begin{align*}
& \frac{\partial \mathcal{E}}{\partial \vec M_{u,J}} =  \int_\mathcal{T} dt \left[ \langle\vec a_z^\theta (\vec x)\tr \vec a_z^\theta (\vec x) \rangle_{q_x} \right]_{[i,i]} \vec M_{u,J}  \\
& \qquad + \int_\mathcal{T} dt \left[\left\langle \nabla_x \vec a_z^\theta (\vec x)\right\rangle_{q_x} \right]_{[:,i]}\tr  \vec S_x \vec A\tr \\
& \qquad  - \int_\mathcal{T} dt \left[\left\langle \vec a_z^\theta (\vec x)\right\rangle_{q_x}\right]_{[:,i]} \tr ( -  \vec A\vec m_x + \vec b) \tr
\end{align*}

Putting all terms together, we obtain the update
\begin{align*}
    \vec M_{u,J} & = \vec B_{1} \inv \left(\vec B_{2} -\vec B_{3} \right)
\end{align*}
with
\begin{align*}
    \vec B_{1} & = \left(\tilde{\vec \Omega} + \int_\mathcal{T} dt \left[ \langle\vec a_z^\theta (\vec x)\tr \vec a_z^\theta (\vec x) \rangle_{q_x} \right]_{[uj,uj]} \right) \\
    \vec B_{2} & =    \int_\mathcal{T} dt  \left[\left\langle \vec a_z^\theta (\vec x)\right\rangle_{q_x}\right]_{[:,uj]} \tr \langle \vec f_q \rangle_{q_x}\tr \\
     \vec B_{3} & =    \int_\mathcal{T} dt \left[\left\langle \nabla_x \vec a_z^\theta (\vec x)\right\rangle_{q_x} \right]_{[:,uj]} \tr \vec S_x \vec A \tr
\end{align*}
and we have defined an indexing operation where $[X]_{[uj,uj]}$ selects the first $M\times M$ and last $LK \times LK$ block of $X$ and $[X]_{[:,uj]}$ selects the first $M$ and last $LK$ columns of $X$. Hence, this selects the appropriate block matrices for the updates. The one-dimensional integrals can be computed efficiently using Gauss-Legendre quadrature.

\subsection{Sparse Gaussian Process dynamics}
Similarly, closed form updates are available in the simpler case, when $\vec f$ is modelled by a classic sparse Gaussian Process, i.e. using inducing points without the additional conditioning on fixed points and Jacobians.
\begin{align*}
	 \vec S_u^k  & =	\vec K_{zz}  \left( \vec K_{zz} + \int_\mathcal{T} dt \left\langle \vec \kappa(\vec Z, \vec x)  \vec \kappa(\vec x, \vec Z) \right\rangle_{q_x} \right) \inv \vec K_{zz} \\
	 	\vec M_u & = \vec S_u^k \vec K_{zz}\inv  \left( \int_\mathcal{T} dt \vec \Phi_1 \vec f_q\tr -  \int_\mathcal{T} dt\vec \Phi_{d1} \vec S_x \vec A\tr \right)
\end{align*}
Where $\vec \Phi_1  =  \left\langle k(\vec x, \vec Z) \right\rangle_{q_x} $ and $\vec \Phi_{d1}  =\left\langle\frac{\partial}{\partial \vec x} k(\vec x, \vec Z)\right\rangle_{q_x}$. 
 
\subsection{Linear dynamics}
Our modelling framework also easily extends to other parameterisation of $\vec f$. For example, in a continuous-time linear dynamical system with $\vec f(\vec x) = -\tilde{\vec A} \vec x + \tilde{\vec b}$ direct minimisation of the KL-divergence between the continuous-time processes leads to the closed form updates
\begin{align*}
\tilde{\vec A}   &= \left( \int_{\mathcal{T}} dt \left(\vec b \langle \vec x \rangle\tr  - \langle \vec f_q (\vec x ) \vec x \tr  \rangle \right) \right) \left( \int_{\mathcal{T}} dt \langle \vec x \vec x \tr  \rangle \right)\inv \\
\tilde{\vec b} & = \frac{1}{T} \int_{\mathcal{T}} dt \left(  \langle \vec f_q(\vec x) \rangle + \vec A \langle \vec x \rangle \right)
\end{align*}
reminiscent of the update equations for the generative parameters of a discrete-time Linear Dynamical System.

\subsection{Output mapping}
We consider an observation model of the form
\begin{equation*}
    \vec y (t_i) = \vec C \vec x(t_i) + \vec d + \vec \epsilon_i
\end{equation*}
where $\vec \epsilon_i \sim \mathcal{N}(\epsilon| 0, \Gamma)$. Dropping all terms that are constant in $\vec C, \vec d$ from the expression for the variational free energy, we have
\begin{align*}
    \mathcal{F^*} &= - \frac{1}{2} \sum_t\left\langle \left( \vec y_t - \vec C \vec x_{t} -\vec d\right) \tr \Gamma\inv\left( \vec y_t - \vec C \vec x_{t} - \vec d\right)   \right\rangle_{q_x}
\end{align*}
Differentiating and setting to zero gives
\begin{align*}
     \vec C^{new}  & =  \left( \sum_t (\vec y_t - \vec d)\vec m_t \tr \right) \left( \sum_t (\vec S_{x,t} + \vec m_{x,t} \vec m_{x,t} \tr) \right) \inv \\
     \vec d^{new} & = \frac{1}{T}\sum_t \left( \vec y_t - \vec C^{new} \vec m_{x,t} \right)
\end{align*}

\section{Chemical reaction dynamics}
The dynamical system used to generate the data in section \ref{sec:exp_chem} is of the form
\begin{align*}
\frac{d \lfloor \mathrm{I}^-\rfloor_A}{dt} & =  \left( k_a \lfloor \mathrm{I}^-\rfloor_A  + k_b \lfloor \mathrm{I}^-\rfloor_A^2\right)\left( S_0 - \lfloor \mathrm{I}^-\rfloor_A \right) \\
& \quad + \frac{F_1  \lfloor \mathrm{I}^-\rfloor_0 }{V_A} -  \frac{(F_3 + F_4)  \lfloor \mathrm{I}^-\rfloor_A }{V_A} +  \frac{F_4 \lfloor \mathrm{I}^-\rfloor_D }{V_A} \\
\frac{d \lfloor \mathrm{I}^-\rfloor_D}{dt} & = \left( k_a \lfloor \mathrm{I}^-\rfloor_D  + k_b \lfloor \mathrm{I}^-\rfloor_D^2\right)\left( S_0 - \lfloor \mathrm{I}^-\rfloor_D \right) \\
& \quad + \frac{F_4 \lfloor \mathrm{I}^-\rfloor_A }{V_D} -\frac{F_4 \lfloor \mathrm{I}^-\rfloor_D }{V_D}
\end{align*}
To generate the simulations, we use the parameter settings
\begin{align*}
    && \lfloor \mathrm{I}^-\rfloor_0 & = \num{4.4e-5} &  k_0 & = \num{2.7e-3}\\
    && V_A & = \num{4e+1} &  F_4 & = \num{3.25e-3} \\
    && V_D & = 1 & F_3 & = k_0 V_a\\
    && k_a & = \num{2.1425e-1} & F_1 & = \half F_3\\
    && k_b & = \num{2.1425e+4} & F_2 & = \half F_3\\
    && S_0 & = \half \left(\lfloor \mathrm{I}^-\rfloor_0 + \num{1.42e-3}\right)\\
\end{align*}

\end{document}